\documentclass[lettersize,journal]{IEEEtran}

\usepackage[switch]{lineno}

\usepackage{microtype}
\usepackage{graphicx}
\usepackage{subcaption}
\usepackage{booktabs} 

\usepackage{diagbox}
\usepackage{makecell}
\usepackage{multirow}
\usepackage{colortbl}
\usepackage{enumitem}

\usepackage{amsmath,amsfonts,amsthm}

\usepackage{algorithmic}
\usepackage{algorithm}
\usepackage{array}
\usepackage[caption=false,font=normalsize,labelfont=sf,textfont=sf]{subfig}
\usepackage{textcomp}
\usepackage{stfloats}
\usepackage{url}
\usepackage{verbatim}
\usepackage{graphicx}
\usepackage{cite}
\usepackage{arydshln}
\usepackage{multirow}
\usepackage[normalem]{ulem}

\usepackage{diagbox}

\usepackage{xcolor}
\usepackage{enumitem}
\usepackage{xspace}
\newcommand{\model}{{D$^2$MoE}\xspace}

\newcommand{\eg}{e.g.,\xspace}
\newcommand{\ie}{i.e.,\xspace}

\theoremstyle{plain}
\newtheorem{theorem}{Theorem}

\theoremstyle{definition}

\theoremstyle{remark}

\usepackage{pdfpages}    
\usepackage{fancyhdr}    

\begin{document}
\title{Learning How Much to Think: Difficulty-Aware Dynamic MoEs for Graph Node Classification}

\author{
Jiajun Zhou, 
Yadong Li,
Xuanze Chen, 
Chen Ma,
Chuang Zhao,
Shanqing Yu, 
Qi Xuan, \IEEEmembership{Senior Member, IEEE}
\thanks{This work was supported in part by National Natural Science Foundation of China (No. 62503423), in part by the National Key Research and Development Program of China (No. 2025YFA1510900), in part by the Key Research and Development Program of Zhejiang Province (No. 2026C02A1233), in part by the Yangtze River Delta Science and Technology Innovation Community Joint Research Project (No. 2026ZY03003, No. 2025CSJGG01000), in part by the Zhejiang Provincial Natural Science Foundation of China (No. LMS25F020005). \emph{(Corresponding authors: Jiajun Zhou.)}}
\thanks{Jiajun Zhou, Yadong Li, Xuanze Chen, Chen Ma, Shanqing Yu and Qi Xuan are with the Institute of Cyberspace Security, Zhejiang University of Technology, Hangzhou 310023, China, with the Binjiang Cyberspace Security Institute of ZJUT, Hangzhou, 310056, China, and with the Soovar Technologies Co., Ltd., Hangzhou 310056, China (e-mail: jjzhou@zjut.edu.cn).}
\thanks{Chuang Zhao is with the Department of Electronic and Computer Engineering, Hong Kong University of Science and Technology, Hong Kong, China.}
}

\markboth{Journal of \LaTeX\ Class Files,~Vol.~14, No.~8, August~2021}%
{Shell \MakeLowercase{\textit{et al.}}: A Sample Article Using IEEEtran.cls for IEEE Journals}


\maketitle



\begin{abstract}
  Mixture-of-Experts (MoE) architectures offer a scalable path for Graph Neural Networks (GNNs) in node classification tasks but typically rely on static and rigid routing strategies that enforce a uniform expert budget or coarse-grained expert toggles on all nodes. This limitation overlooks the varying discriminative difficulty of nodes and leads to under-fitting for hard nodes and redundant computation for easy ones. To resolve this issue, we propose \model, a novel framework that shifts the focus from static expert selection to node-wise expert resource allocation. By using predictive entropy as a real-time proxy for difficulty, \model employs a difficulty-driven top-$p$ routing mechanism to adaptively concentrate expert resources on hard nodes while reducing overhead for easy ones, achieving continuous and fine-grained expert budget scaling for node classification. Experiments on 13 benchmarks demonstrate that \model achieves consistent state-of-the-art performance, surpassing leading baselines by up to 7.92\% in accuracy on heterophilous graphs. Notably, on large-scale graphs, it reduces memory consumption by up to 73.07\% and training time by 46.53\% compared to the best-performing Graph MoE, thereby validating its superior efficiency. 
\end{abstract}
    
\begin{IEEEkeywords}
    Graph Neural Networks, Mixture of Experts, Node Classification
\end{IEEEkeywords}

\section{Introduction}\label{sec:intro}
\IEEEPARstart{G}{raph} Neural Networks (GNNs) have emerged as the leading paradigm for representation learning on graph-structured data. By iteratively aggregating neighborhood information via message passing, GNNs have achieved remarkable success in homophily-dominant scenarios where connected nodes tend to share similar features and labels. However, as real-world applications scale to massive datasets with increasingly complex patterns, a demand for higher model reasoning capacity arises. To capture intricate structural dependencies and rich semantic information, the research community have introduced a broad spectrum of advanced architectural mechanisms. These encompass high-order message passing~\cite{H2GCN} to expand receptive fields, multi-channel spectral filtering~\cite{FAGCN,luan2022revisiting} to adaptively capture varying frequency signals, and the recent proliferation of Graph Transformers (GTs)~\cite{NAGphormer,ASN-GT,DIFFormer,Exphormer} designed to model global, all-pair node interactions.

While these innovations have successfully pushed the boundaries of model capacity, they simultaneously introduce a critical dilemma characterized by two intertwined challenges: 
1) \textbf{Redundant computational overhead in dense paradigms.} The conventional dense paradigm dictates that every node must engage the entire model capacity during inference, regardless of its discriminative difficulty. This indiscriminate computational overhead significantly hinders the scalable deployment of large-capacity models on massive graphs.
2) \textbf{Rigid architectural priors versus node heterogeneity.} Applying a uniform architectural prior globally intrinsically lacks the flexibility to adapt to diverse local graph patterns, fundamentally limiting the model's ability to generalize across varying structural contexts.
To reconcile the conflict between expressive capacity and computational efficiency, the Mixture-of-Experts (MoE) architecture~\cite{shazeer2017outrageously} has been increasingly adopted in the graph domain. By conditionally activating only a sparse subset of expert networks, MoEs offer a pathway to scale model expressiveness without incurring a proportional surge in computational cost. 
However, the routing paradigms of existing Graph MoEs remain fundamentally static and rigid. They typically enforce uniform routing strategies, manifesting either as fixed graph expert budgets across all nodes like GMoE~\cite{GMoE} or relying on coarse-grained, heuristic binary toggles like Mowst~\cite{Mowst}. Consequently, \textbf{\emph{this inherent rigidity disregards the naturally varying discriminative difficulty and heterogeneous reasoning capacity demands of individual nodes.}}

This static design fundamentally contradicts the pervasive \textit{heterogeneity} characterizing real-world graph data. As corroborated by our motivational experiments detailed in \textbf{Section~\ref{sec: obs}}, real-world graphs comprise diverse node patterns with varying levels of discriminative difficulty. On one hand, ``easy'' nodes, typically residing in homophilous regions with consistent neighborhood signals, can be accurately classified with minimal computation. Forcing such nodes to be processed through a wide array of multiple experts inevitably leads to \textit{over-computation} and the potential introduction of irrelevant noise from extraneous expert branches. On the other hand, ``hard'' nodes are frequently located at heterophilous boundaries or within densely entangled semantic regions, demanding higher non-linearity and a larger reasoning capacity to effectively disentangle conflicting neighborhood information. For these challenging nodes, imposing a fixed, small expert budget ($k$) severely constrains the model's expressiveness, inevitably resulting in \textit{under-fitting}. Consequently, this static ``one-size-fits-all'' routing paradigm creates a profound resource mismatch: it caps the performance ceiling on hard nodes while simultaneously squandering computational resources on simple ones.

Recently, dynamic routing mechanisms that adapt expert budgets based on instance difficulty have shown remarkable success in Large Language Models (LLMs) and Vision Transformers~\cite{huang2024harder}. However, directly porting these sequence-level routing techniques to Graph Neural Networks (GNNs) presents unique challenges. First, unlike independent tokens in LLMs, nodes in a graph are topologically entangled (non-i.i.d.). The discriminative difficulty of a node stems not only from its intrinsic feature ambiguity but also from its structural surroundings, such as neighborhood heterophily and topological noise. Second, the standard LLM practice of calculating routing weights on-the-fly based on intermediate hidden states can destabilize the recursive message-passing process in GNNs.

To bridge this gap and tailor dynamic MoEs for graph topology, we propose \textbf{D}ifficulty-Aware \textbf{D}ynamic \textbf{M}ixture-\textbf{o}f-\textbf{E}xperts (\model), a novel framework that shifts the routing paradigm from rigid, static expert selection to \textit{continuous, fine-grained expert budget scaling}. Specifically, \model leverages predictive entropy as a real-time proxy for discriminative difficulty. By dynamically adjusting a cumulative routing probability threshold $p$ tailored to each node, \model employs an adaptive top-$p$ routing mechanism to dictate sparse, on-demand expert activation. By explicitly driving expert resource allocation via predictive uncertainty, \model successfully reconciles the aforementioned dilemma, achieving a trade-off between instance-wise reasoning accuracy and system-level computational efficiency on large-scale graphs.
Our main contributions are summarized as follows:
\begin{itemize}[leftmargin=10pt, nosep]
    \item \textbf{New Paradigm:} We propose \model, a novel framework that transcends rigid static routing by enabling \textit{continuous and fine-grained expert budget scaling}, resolving the resource mismatch dilemma by aligning allocation with instance-wise discriminative difficulty.
    \item \textbf{Novel Methodology:} We design a \textit{difficulty-aware top-$p$ routing} that employs predictive entropy as a real-time difficulty proxy. This strategy uses dense expert ensembles for hard nodes while enforcing sparsity for easy ones.
    \item \textbf{Superior Performance:} Extensive experiments on 13 datasets confirm that \model not only achieves SOTA performance for node classification but also exhibits remarkable efficiency, reducing memory usage and training time by up to 73.07\% and 46.53\%, respectively.
\end{itemize}

\section{Preliminaries and Related Works}
\label{sec:preliminaries}

\subsection{Notation and Problem Definition}
Let $\mathcal{G} = (\mathcal{V}, \mathcal{E})$ denote a graph, where $\mathcal{V} = \{v_1, \dots, v_N\}$ is the set of $N$ nodes and $\mathcal{E}$ represents the set of edges. The topological structure is described by the adjacency matrix $\boldsymbol{A} \in \{0, 1\}^{N \times N}$, where $\boldsymbol{A}_{ij} = 1$ if an edge exists between node $v_i$ and $v_j$, and $\boldsymbol{A}_{ij} = 0$ otherwise. We denote the neighborhood set of node $v_i$ as $\mathcal{N}(v_i) = \{v_j \in \mathcal{V} \mid \boldsymbol{A}_{ij} = 1\}$, which includes the node itself if self-loops exist.
Each node is associated with a feature vector $\boldsymbol{x}_v \in \mathbb{R}^d$, forming the feature matrix $\boldsymbol{X} \in \mathbb{R}^{N \times d}$, where $d$ is the dimension of the input features. The node label information is represented by a matrix $\boldsymbol{Y} \in \{0, 1\}^{N \times C}$, where $C$ is the number of classes. For a node $v_i$ belonging to class $c$, the row vector $\boldsymbol{y}_i$ is a one-hot vector with the $c$-th element being 1. Given a subset of labeled nodes $\mathcal{V}_L \subset \mathcal{V}$, the semi-supervised node classification task aims to learn a function $f: \mathcal{V} \to \mathbb{R}^C$ to predict the labels of the remaining unlabeled nodes $\mathcal{V}_U = \mathcal{V} \setminus \mathcal{V}_L$ by minimizing the prediction error on $\mathcal{V}_L$.

\subsection{Graph Neural Networks}
GNNs follow the message-passing paradigm. For a generic GNN layer $\ell+1$, the representation $\boldsymbol{h}_v^{(\ell+1)}$ of node $v$ is updated by aggregating messages from its neighbors $\mathcal{N}(v)$:
\begin{equation}
    \boldsymbol{h}_v^{(\ell+1)} = \operatorname{UP}\left(\boldsymbol{h}_v^{(\ell)}, \operatorname{AGG}\left(\left\{\boldsymbol{h}_u^{(\ell)} \mid u \in \mathcal{N}(v)\right\}\right)\right),
\end{equation}
where $\boldsymbol{h}_v^{(0)} = \boldsymbol{x}_v$.
$\operatorname{AGG}(\cdot)$ collects messages from neighbors, while $\operatorname{UP}(\cdot)$ combines the aggregated message with the node's current state. 
This mechanism enables GNNs to capture local structural patterns but typically applies shared parameters across all nodes, enforcing a uniform architectural prior.

\subsection{Mixture of Experts with Its Applications to Graphs}
\subsubsection{General MoE Mechanism}
Mixture-of-Experts (MoE) \cite{shazeer2017outrageously} scales model capacity by using a set of $K$ expert networks $\mathcal{F} = \{f_1, \cdots, f_K\}$ alongside a router $\operatorname{R}(\cdot)$. For an input $\boldsymbol{h}$, the router first computes a routing distribution $\boldsymbol{\pi} =\operatorname{R}(\boldsymbol{h}) \in \mathbb{R}^K$ across all experts.
The top-$k$ routing then selects the $k$ experts with the highest scores in $\boldsymbol{\pi}$. 
The final sparse routing weights are obtained by normalizing the scores of the selected experts while masking others to zero.
Finally, the model output is computed as the weighted summation of the activated experts.

\subsubsection{Adoption in GNNs} 
Inspired by the success of MoE in LLMs, recent studies have introduced this paradigm to GNNs to tackle heterogeneity. GMoE~\cite{GMoE} integrates experts with varying aggregation hops, enabling the dynamic selection of optimal receptive fields to capture multi-scale structural information. With a similar focus on structural variability, DAMoE~\cite{DAMoE} addresses the depth sensitivity issue arising from diverse graph scales. 
NodeMoE~\cite{NodeMoE} employs a bank of graph filters (\eg low-pass, high-pass) as experts, selecting the filter that best aligns with the local homophily pattern of each node.
GNNMoE~\cite{GNNMoE} decouples message passing into atomic operations (\ie propagation and transformation) to construct diverse expert networks.
Mowst~\cite{Mowst} adopts a ``strong-weak'' strategy, utilizing an MLP-based confidence metric to conditionally activate GNN experts only for low-confidence nodes. 
Moscat~\cite{Moscat} introduces a decoupled expert-gating paradigm that independently trains GNNs of varying depths as scope experts and dynamically combines their predictions at test time to enhance generalization
Despite effectively utilizing diverse experts to handle node heterogeneity, these methods remain constrained by rigid expert resource allocation mechanisms. Most approaches adhere to static routing with a fixed expert budget, while others are limited to coarse-grained expert toggles. Crucially, they lack the flexibility for \textit{continuous, fine-grained expert budget scaling}, inevitably leading to resource mismatch where hard nodes suffer from under-fitting while easy nodes incur computational waste.

\begin{figure}[!htb]
  \centering
  \includegraphics[width=\linewidth]{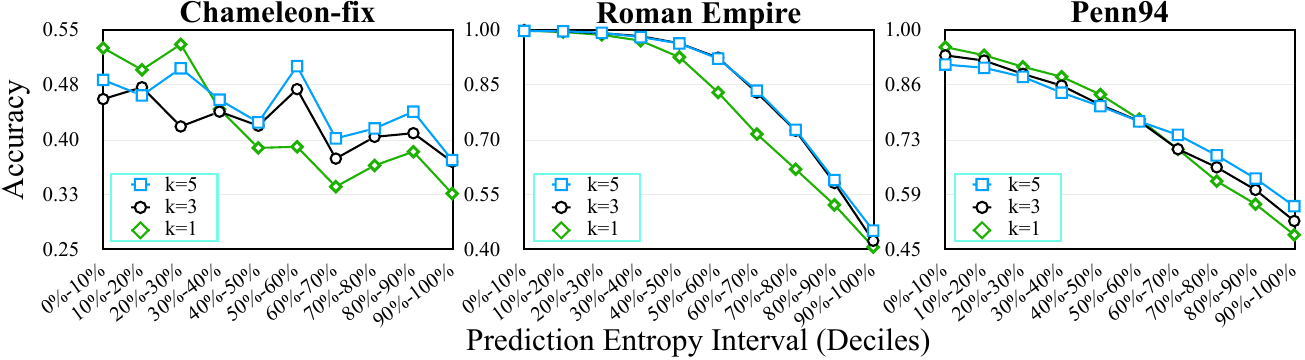}
  \caption{Impact of static expert budgets (top-$k$) on node classification performance across predictive entropy intervals.}
  \label{fig: obs-main}
\end{figure}

\section{Motivation and Theoretical Analysis}\label{sec: obs_theorem}
\subsection{Analysis of Motivational Experiments}\label{sec: obs}

To investigate the inductive bias and limitations of static routing strategies (\eg top-$k$) in existing Graph MoEs, we conduct a comprehensive controlled motivational experiment. We construct a vanilla sparse Graph MoE framework comprising a top-$k$ router and a bank of $K=6$ independent experts, where each expert is instantiated as a standard 2-layer GCN. To analyze the impact of the computational budget, we evaluate the node classification performance of three variants with fixed static routing ($k \in \{1, 3, 5\}$) across nodes stratified by their discriminative difficulty. Crucially, to ensure a consistent difficulty baseline and prevent ``population drift'' caused by varying model capacities, we employ a fixed, independent proxy model to quantify node discriminative difficulty. Specifically, we utilize a fully converged 2-layer GCN as a ``teacher'' to calculate the predictive entropy for all test nodes. Based on this fixed predictive entropy distribution, test nodes are stratified into immutable difficulty intervals (deciles from 0\% to 100\%), ensuring that all expert variants are strictly evaluated on identical subsets of ``easy'' and ``hard'' nodes.

As illustrated in Figure~\ref{fig: obs-main}, the comprehensive results across diverse datasets validate a fundamental \textit{expert resource mismatch} inherent in static routing. Regardless of the graph topology, two distinct and consistent patterns emerge:

\textbf{1) Capacity Saturation in Low-Entropy Regions:} In intervals characterized by low predictive entropy (\eg $\mathcal{U}_v \in [0\%, 20\%]$), the performance curves of models with varying expert budgets exhibit minimal variance and frequently overlap. Notably, the single-expert baseline ($k=1$) matches or even marginally exceeds the accuracy of multi-expert ensembles. This saturation explicitly suggests that ``easy'' nodes, typically residing in homophilous regions with consistent structural signals, are fully adequately supervised by a minimal expert budget. Consequently, enforcing a larger $k$ in these regions incurs unnecessary computational overhead and risks introducing noise from less relevant experts.

\textbf{2) Capacity Craving in High-Entropy Regions:} Conversely, a substantial performance divergence manifests as predictive entropy increases (\eg $\mathcal{U}_v > 0.5$). For these challenging nodes, often located at heterophilous boundaries or noisy neighborhoods, ensembles with larger budgets ($k=3, 5$) demonstrate significantly superior performance compared to the single-expert variant. This trend confirms that a fixed, small budget inevitably leads to under-fitting for ``hard'' nodes, as they inherently demand the activation of more experts to reduce variance and collaboratively resolve semantic ambiguity.

Ultimately, these observations reveal that static routing paradigms result in an inescapable dilemma: over-computation for easy nodes and simultaneous under-fitting for hard ones. This profound resource mismatch underscores the critical necessity for a difficulty-driven adaptive routing paradigm, directly motivating the design of our \model framework.

\vspace{-5pt}
\subsection{Theoretical Justification}\label{sec:theory}

To theoretically justify the necessity of the dynamic computational allocation mechanism in $D^2$MoE, we analyze the generalization error of the Mixture-of-Experts (MoE) architecture in node classification tasks through the lens of the Bias-Variance Trade-off.
Given a node $v$ with predictive entropy $\mathcal{U}_v$, the expected Mean Squared Error $\mathcal{L}(v, k)$ of its Top-$k$ expert ensemble can be decomposed into bias, variance, and irreducible noise: $\mathcal{L}(v, k) = \text{Bias}^2(k) + \text{Var}(k) + \epsilon$. In the context of dynamically weighted sparse MoEs for graph neural networks, we establish the following two fundamental assumptions that reflect realistic physical settings:

\textbf{Assumption 1 (Weighted Bias Accumulation).} 
In a sparse MoE, the router outputs normalized routing weights $\sum_{i=1}^k \pi_i = 1$, and the experts are sorted in descending order of confidence ($\pi_1 \ge \pi_2 \ge \dots \ge \pi_k$). Expanding the number of activated experts (increasing $k$) implies introducing relatively lower-performing tail experts. Due to the zero-sum nature of the weights, introducing tail experts inevitably siphons probability mass away from the optimal top experts. Although the soft weight allocation mechanism of Softmax effectively mitigates the severe bias dilution caused by uniform averaging, the bias still strictly and monotonically increases with $k$. We model this sub-linear bias growth via a generalized polynomial:
\begin{equation}
    \text{Bias}^2(k) \propto \beta k^\mu \quad (\beta > 0, \mu > 0)
\end{equation}

\textbf{Assumption 2 (Uncertainty-Driven Reducible Variance).} 
The predictive entropy $\mathcal{U}_v$ of node $v$ reflects the model's uncertainty regarding its classification, which directly corresponds to a high initial prediction variance. In practical ensembles, a certain degree of positive correlation ($\rho > 0$) typically exists among experts, implying a theoretical lower bound for the variance. However, increasing the number of experts $k$ can still monotonically decrease the \textit{reducible} portion of the variance. We model this uncertainty-driven variance reduction as:
\begin{equation}
    \text{Var}(k) \propto \rho \sigma^2 + \frac{\alpha \mathcal{U}_v}{k^\varphi  } \quad (\alpha > 0, \varphi  > 0)
\end{equation}
where $\sigma^2$ denotes the variance of a single expert’s prediction. By minimizing the generalized error $\mathcal{L}(v, k)$, we can derive the analytical relationship between node discriminative difficulty and the optimal expert budget (proof in Appendix~\ref{app:detailed_proof}):
\begin{theorem}[\textbf{Generalized Uncertainty-Driven Scaling Law}]
  \label{thm:scaling_law}
  Under Assumptions 1 and 2, for any node $v$, there exists a positive monotonic relationship between its predictive entropy $\mathcal{U}_v$ and the optimal number of active experts $k^*$ that minimizes its generalization error:
  \begin{equation}
      k^*(v) \propto \left(\mathcal{U}_v\right)^{\frac{1}{\mu+\varphi }}
  \end{equation}
\end{theorem}

\textit{Remark.}
Theorem \ref{thm:scaling_law} provides a rigorous mathematical foundation for \model to abandon static Top-$k$ routing. The theorem proves that in realistic scenarios accounting for expert weighting and correlation, the system must establish a \textbf{dynamic, non-linear positive mapping} between task uncertainty and computational capacity to achieve optimal generalization. 
This serves as the core motivation for the routing mechanism design in Section \ref{sec: method} (Eq. (\ref{eq: pv})). While the theoretical scaling law exhibits an unbounded power-law growth, a physical MoE system has a strictly bounded total expert capacity. Therefore, we employ a Sigmoid function as a continuous threshold mapper. The Sigmoid function preserves the non-linear, monotonically increasing property dictated by Theorem~\ref{thm:scaling_law}, while naturally mapping the unbounded predictive entropy into a valid probability simplex $(0, 1)$ to determine the adaptive expert budget. This design serves as a rigorous practical instantiation of the theoretical scaling law under realistic hardware constraints.

\section{Methodology}\label{sec: method}
\begin{figure}[!htb]
  \centering
  \includegraphics[width=0.95\linewidth]{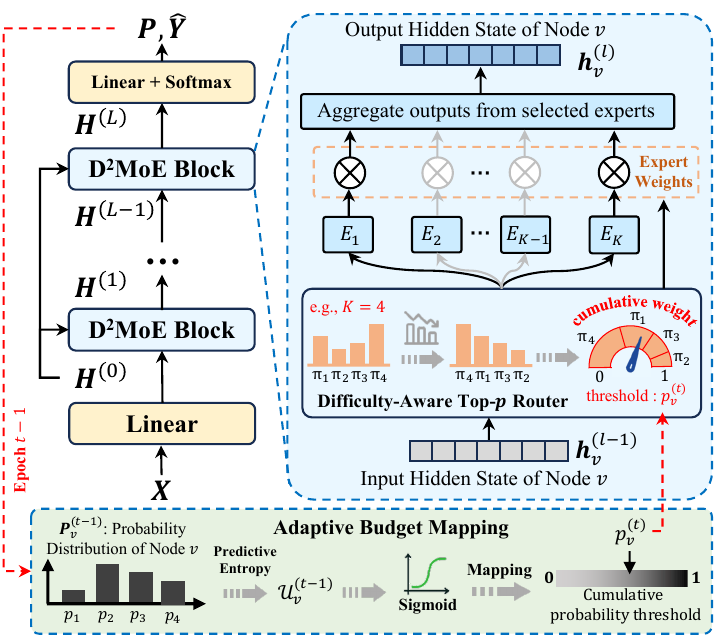}
  \caption{The architecture of \model. The complete workflow proceeds as follows: 
  (1) \model first quantifies node-wise difficulty via predictive entropy;
  (2) Top-$p$ router adaptively scales the expert resources, activating a sparse set of experts for easy nodes while mobilizing a comprehensive ensemble for hard nodes; 
  (3) Selected expert outputs are fused to update node representations, which generate predictions and initiate the next difficulty quantification.}
  \label{fig: framework}
\end{figure}

Based on the motivational analysis in Section~\ref{sec: obs_theorem}, we propose Difficulty-Aware Dynamic Mixture-of-Experts (\model) for graph node classification, a novel framework that transforms node discriminative difficulty into a node-wise computational budget. \model enables a dynamic routing paradigm where the model adaptively decides \textit{how many} experts are required for each node during inference, rather than relying on a static routing configuration. The overall architecture of \model is illustrated in Figure~\ref{fig: framework}.

\subsection{Backbone Architecture of \model}\label{sec:backbone}
The overall architecture of \model follows a stacked layer-wise Mixture-of-Experts paradigm, designed to iteratively refine node representations via dynamic expert interaction.

\subsubsection{Initial Embedding} 
First, the input node feature matrix $\boldsymbol{X} \in \mathbb{R}^{N \times d}$ is projected into a latent space compatible with the experts. We employ a linear transformation parameterized by $\boldsymbol{W}_0 \in \mathbb{R}^{d \times h}$, followed by a ReLU activation and a dropout operation to obtain the initial node embeddings:
\begin{equation}
  \label{eq: initial-pro}
    \boldsymbol{H}^{(0)} = \operatorname{Dropout}(\operatorname{ReLU}(\boldsymbol{X}\boldsymbol{W}_0)),
\end{equation}
where $\boldsymbol{H}^{(0)} \in \mathbb{R}^{N \times h}$ and $h$ is the hidden dimension.

\subsubsection{Stacked \model Layers}
Subsequently, $\boldsymbol{H}^{(0)}$ serves as the input to the core network composed of $L$ stacked \model blocks. 
In each block $l \in \{1, \ldots, L\}$, the input $\boldsymbol{H}^{(l-1)}$ is processed in parallel by a difficulty-aware router and a bank of experts $\mathcal{F}^{(l)} = \{f_1^{(l)}, \dots, f_K^{(l)}\}$.
Each expert $f_i^{(l)}$ is instantiated as a standard GNN layer (\eg GCN or GraphSAGE) with independent parameters, allowing them to differentiate and specialize in capturing diverse local patterns during training. The output of the $i$-th expert is: 
\begin{equation}
    \boldsymbol{Z}_i^{(l)} = f_i^{(l)}(\boldsymbol{H}^{(l-1)}, \boldsymbol{A}).
\end{equation}

\subsubsection{Difficulty-Aware Dynamic Routing}
Simultaneously, the router $\operatorname{R}^{(l)}(\cdot)$, instantiated as a Multi-Layer Perceptron (MLP), generates routing scores based on the input features. It maps the node embeddings to the expert score dimension:
\begin{equation}
  \begin{aligned}
    \boldsymbol{\pi}_v^{(l)} &= \operatorname{Softmax}(\operatorname{R}^{(l)}(\boldsymbol{h}_v^{(l-1)}))\\
     &= \operatorname{Softmax}(\boldsymbol{W}_2 \cdot \sigma(\boldsymbol{W}_1 \boldsymbol{h}_v^{(l-1)})),
  \end{aligned}
\end{equation}
where $\boldsymbol{W}_1$ and $\boldsymbol{W}_2$ are learnable weights, $\sigma$ denotes the activation function, and $\boldsymbol{\pi}_v^{(l)} \in \mathbb{R}^K$ represents the probability distribution over experts for node $v$ prior to selection.

Unlike traditional rigid static routing, \model employs a difficulty-aware dynamic routing mechanism that utilizes $\boldsymbol{\pi}_v^{(l)}$ to adaptively calculate a threshold and select a variable set of the most relevant experts $\widehat{\mathcal{F}}_v^{(l)}$ ({detailed in Section \ref{sec:routing}}). To effectively aggregate information from these selected experts while mitigating the over-smoothing problem common in deep GNNs, we integrate residual connections. The update rule for node $v$ at block $l$ is formulated as:
\begin{equation}
    \boldsymbol{h}_v^{(l)} = \boldsymbol{h}_v^{(l-1)} + \sum_{i \in \widehat{\mathcal{F}}_v^{(l)}} \bar{\pi}_{v,i}^{(l)} \cdot [\boldsymbol{Z}_i^{(l)}]_v,
\end{equation}
where $[\boldsymbol{Z}_i^{(l)}]_v$ is the output of expert $f_i^{(l)}$ for node $v$. The re-normalized routing weight $\bar{\pi}_{v,i}^{(l)}$ ensures numerical stability and that the contribution of active experts sums to 1:
\begin{equation}
    \bar{\pi}_{v,i}^{(l)} = \begin{cases}
    \frac{{\pi}_{v,i}^{(l)}}{\sum_{j \in \widehat{\mathcal{F}}_v^{(l)}} {\pi}_{v,j}^{(l)}}, & \text{if } i \in \widehat{\mathcal{F}}_v^{(l)}, \\
    0, & \text{otherwise}.
    \end{cases}
\end{equation}
\subsubsection{Output and Prediction}
Finally, the aggregated representation undergoes normalization and non-linear activation to produce the layer output:
\begin{equation}
    \boldsymbol{H}^{(l)} = \operatorname{Dropout}(\operatorname{ReLU}(\operatorname{Norm}(\boldsymbol{H}^{(l)}))),
\end{equation}
where $\operatorname{Norm}(\cdot)$ denotes a normalization layer.
After $L$ layers, the final node representation $\boldsymbol{H}^{(L)}$ is fed into a linear classification head to generate the prediction probability distribution ${\boldsymbol{P}}$ and the prediction labels $\hat{\boldsymbol{Y}}$:
\begin{equation}
  \boldsymbol{P} = \operatorname{Softmax}(\operatorname{Linear}(\boldsymbol{H}^{(L)})), \  \hat{\boldsymbol{Y}}=\operatorname{argmax}(\boldsymbol{P}).
\end{equation}


\subsection{Difficulty-Aware Dynamic Routing}
\label{sec:routing}

The core innovation of \model lies in its routing mechanism, which transforms the abstract concept of ``discriminative difficulty'' into a concrete, instance-wise computational budget. This process operates in three phases: difficulty quantification, adaptive budget mapping, and dynamic execution.

\subsubsection{Discriminative Difficulty Quantification}
First, we establish a metric to quantify the discriminative difficulty of each node. We employ predictive entropy as a real-time proxy. To enable a closed-loop bootstrapping strategy, we utilize the model's prediction from the \textit{previous} training epoch $t-1$ to guide the resource allocation at the \textit{current} epoch $t$. For a node $v$, its discriminative difficulty $\mathcal{U}_v^{(t-1)}$ is defined as the Shannon entropy of its predicted class distribution $\boldsymbol{P}_v^{(t-1)}$:
\begin{equation}\label{eq: difficulty-proxy}
    \mathcal{U}_v^{(t-1)} = -\frac{1}{\log C} \sum_{c=1}^{C} P_{v,c}^{(t-1)} \cdot \log P_{v,c}^{(t-1)},
\end{equation}
where $C$ is the number of classes, and the normalization term $\log C$ ensures $\mathcal{U}_v^{(t-1)} \in [0, 1]$. Higher entropy indicates greater uncertainty in the model's decision-making, signaling a ``hard'' node.
To ensure training stability during the cold-start phase (\ie $t=0$), we enforce full activation ($p_v^{(0)}$) for all nodes at initialization, which mandates the utilization of the entire expert ensemble, preventing the router from collapsing into a trivial solution early during training.

\subsubsection{Adaptive Budget Mapping}
With the quantified discriminative difficulty $\mathcal{U}_v^{(t-1)}$, the subsequent challenge is to translate this abstract signal into a concrete computational budget for the current epoch. 
A naive strategy might map $\mathcal{U}_v^{(t-1)}$ linearly to a discrete number of experts. However, this is suboptimal as the correlation between discriminative difficulty and expert demand is inherently non-linear.

In this work, we argue that effective resource allocation fundamentally responds to the demand for decision confidence. Intuitively, for ``easy'' nodes with low $\mathcal{U}_v^{(t-1)}$, signals from the dominant expert suffice to ensure certainty. In contrast, ``hard'' nodes with high $\mathcal{U}_v^{(t-1)}$ necessitate the aggregation of opinions from multiple experts to establish a decision consensus and mitigate prediction risk. Consequently, rather than forcing the model to predict a discrete number of experts, we reformulate the task as determining a continuous \textit{cumulative probability threshold} $p_v^{(t)}$. This threshold represents the necessary ``information coverage'' for reliable prediction on node $v$ during epoch $t$. To capture the complex non-linear correlation between discriminative difficulty and expert demand, we design a Sigmoid-based adaptive mapping mechanism:
\begin{equation}\label{eq: pv}
  p_v^{(t)} = \operatorname{Sigmoid}\left(\gamma \cdot \left(\mathcal{U}_v^{(t-1)} - \bar{\mathcal{U}}^{(t-1)}\right)\right),
\end{equation}
where $\gamma$ acts as a sensitivity coefficient controlling the steepness of the Sigmoid function, and a larger $\gamma$ implies that the model is more sensitive to subtle variations in discriminative difficulty. $\bar{\mathcal{U}}^{(t-1)}$ represents the mean value of predictive entropy across all nodes at epoch $t-1$, serving as a dynamic baseline of discriminative difficulty.
Through this design, a higher $\mathcal{U}_v^{(t-1)}$ naturally translates into a higher threshold $p_v^{(t)}$, which in turn necessitates accumulating weights from more experts in the subsequent routing stage.

\subsubsection{Dynamic Top-$p$ Routing}
Finally, we implement a cumulative probability routing strategy (top-$p$ routing) to replace the rigid static top-$k$ routing mechanism.
Mathematically, our goal is to select the \textit{minimal} set of experts sufficient to satisfy the cumulative probability threshold $p_v^{(t)}$. This can be formulated as a constrained optimization problem.
Given the routing scores $\boldsymbol{\pi}_v^{(l)}\in \mathbb{R}^K$, the active expert set $\widehat{\mathcal{F}}_v^{(l)}$ is obtained by solving:
\begin{equation}\label{eq: top-p solving}
    \widehat{\mathcal{F}}_v^{(l)} = \underset{\mathcal{S} \subseteq \{1, \cdots, K\}}{\arg\min} |\mathcal{S}| \quad \text{s.t.} \quad \sum_{i \in \mathcal{S}} \pi_{v,i}^{(l)} \ge p_v^{(t)}.
\end{equation}
The optimal solution to Eq.~(\ref{eq: top-p solving}) is efficiently obtained by a greedy strategy: selecting experts with the largest routing scores until the cumulative sum exceeds $p_v^{(t)}$.
This mechanism ensures that easy nodes (low $p_v^{(t)}$) are processed by a minimal number of experts (often $|\mathcal{S}|=1$), while hard nodes (high $p_v^{(t)}$) automatically trigger a larger ensemble to resolve ambiguity.

In summary, by transforming posterior predictive entropy signals into prior routing decisions, \model effectively dismantles the constraints of rigid static routing. This mechanism evolves \model into a dynamic ``divide-and-conquer'' system, empowering \model to adaptively learn \textit{how much to think} for each specific instance. Consequently, \model achieves continuous, fine-grained expert resource allocation, maximizing expressiveness for challenging samples while preserving computational efficiency for simpler ones.

\subsection{Training Objectives and Regularization}\label{sec:training}
\model is optimized in an end-to-end manner. We formulate the overall objective function, which combines the primary classification loss with auxiliary regularization terms designed to ensure sparse and balanced expert utilization.

\subsubsection{Classification Objective}
The primary goal is the node classification task, optimized via the standard Cross-Entropy loss. Formally, the task loss is defined as:
\begin{equation}\label{eq: loss-main}
  \mathcal{L}_\text{task} = - \frac{1}{|\mathcal{V}_\text{train}|} \sum_{v \in \mathcal{V}_\text{train}} \boldsymbol{Y}_{v}\cdot\log \boldsymbol{P}_{v},
\end{equation}
where $\mathcal{V}_\text{train}$ represents the set of training nodes, and $\boldsymbol{P}_{v}$ and $\boldsymbol{Y}_{v}$ represent the predicted probability distribution and the one-hot ground-truth label of node $v$, respectively.

\begin{algorithm}[!htb]
  \caption{Training Procedure of \model}
  \label{alg:training}
\begin{algorithmic}[1]
  \STATE {\bfseries Input:} Graph $\mathcal{G}=(\mathcal{V}, \mathcal{E}, \boldsymbol{X}, \boldsymbol{Y})$, Max epochs $T$, Hyperparameters $\gamma, \alpha, \beta$.
  \STATE {\bfseries Output:} Trained model parameters $\boldsymbol{\Theta}$.
  
  \STATE \textbf{Initialization:} Initialize model parameters $\boldsymbol{\Theta}$.
  \STATE \textbf{Cold Start:} Initialize cumulative routing probability threshold $p_v^{(0)} \leftarrow 1$ for all nodes $v \in \mathcal{V}$.
  
  \FOR{epoch $t=1$ to $T$}
      \STATE \textcolor{gray}{// Phase 1: Adaptive Budget Mapping (Section \ref{sec:routing})}
      \STATE Get dynamic baseline $\bar{\mathcal{U}}^{(t-1)} \leftarrow \operatorname{Mean}(\mathcal{U}^{(t-1)})$.
      \FOR{each node $v \in \mathcal{V}$}
          \STATE Compute routing threshold $p_v^{(t)}$ via Eq.~(\ref{eq: pv}).
      \ENDFOR
      
      \STATE \textcolor{gray}{// Phase 2: Forward Propagation with Dynamic Routing}
      \STATE Compute initial embeddings $\boldsymbol{H}^{(0)}$.
      \FOR{layer $l=1$ to $L$}
          \FOR{each node $v \in \mathcal{V}$}
              \STATE Generate expert scores $\boldsymbol{\pi}_v^{(l)}$ via Router.
              \STATE \textbf{Dynamic Top-$p$ Execution:} Select minimal expert set $\widehat{\mathcal{F}}_v^{(l)}$ satisfying cumulative sum $\ge p_v^{(t)}$ via Eq.~(\ref{eq: top-p solving}).
              \STATE Aggregate expert outputs with residual connection.
          \ENDFOR
      \ENDFOR
      \STATE Obtain final predictions $\boldsymbol{P}^{(t)}$.
      
      \STATE \textcolor{gray}{// Phase 3: Optimization}
      \STATE Calculate Classification Loss $\mathcal{L}_\text{task}$ via Eq.~(\ref{eq: loss-main}).
      \STATE Calculate Regularization $\mathcal{L}_\text{RE}$ and $\mathcal{L}_\text{LB}$ via Eq.~(\ref{eq: loss-LB}).
      \STATE Update $\boldsymbol{\Theta}$ by minimizing $\mathcal{L} = \mathcal{L}_\text{task} + \lambda_1 \mathcal{L}_\text{RE} + \lambda_2 \mathcal{L}_\text{LB}$.
      
      \STATE \textcolor{gray}{// Phase 4: Proxy Update (Bootstrap)}
      \FOR{each node $v \in \mathcal{V}$}
          \STATE Update difficulty proxy $\mathcal{U}_v^{(t)}$ using current prediction $\boldsymbol{P}_v^{(t)}$ via Eq.~(\ref{eq: difficulty-proxy}).
      \ENDFOR
  \ENDFOR
  \STATE \textbf{return} Optimized parameters $\boldsymbol{\Theta}$.
\end{algorithmic}
\end{algorithm}

\subsubsection{Regularization}
While the difficulty-aware mechanism allows for flexible resource allocation, unconstrained optimization in MoEs is often prone to routing collapse. We identify two critical risks: 
(1) \textit{Loss of Sparsity.} To minimize prediction risk, the router tends to assign uniform weights to all experts. Crucially, in our top-$p$ routing, a flat (high-entropy) distribution forces the activation of numerous experts to meet the cumulative threshold $p_v$, effectively reverting the model to a computationally expensive dense network;
(2) \textit{Load Imbalance.} The router may converge to utilizing only a small subset of strong experts for all nodes, leaving others undertrained and wasting model capacity.
To address these issues, we impose the following constraints:

\begin{itemize}[leftmargin=10pt, nosep]
  \item \textbf{Sparsity via Routing Entropy Minimization.}
  To prevent the router from degenerating into dense activation patterns, we aim to minimize the uncertainty of the routing distribution. By minimizing the entropy of routing distribution $\boldsymbol{\pi}_v^{(l)}$, we encourage the probability mass to concentrate on a small subset of dominant experts. Mathematically, a lower entropy corresponds to a sharper distribution, which enables the cumulative probability threshold $p_v$ to be satisfied with fewer experts (\ie a smaller $|\mathcal{S}|$), thereby enhancing sparsity. The routing entropy loss is formulated as:
  \begin{equation}\label{eq: loss-RE}
      \mathcal{L}_\text{RE} = - \frac{1}{N \cdot L} \sum_{l=1}^{L} \sum_{v=1}^{N} \sum_{i=1}^{K} \pi_{v,i}^{(l)} \cdot \log \pi_{v,i}^{(l)}.
  \end{equation}
  \item \textbf{Utilization via Load Balancing.}
  To prevent expert collapse where certain experts are over-utilized while others remain idle, we enforce balanced expert utilization across all nodes via the load balancing loss~\cite{shazeer2017outrageously}:
  \begin{equation}\label{eq: loss-LB}
      \mathcal{L}_\text{LB} = \sum_{l=1}^{L} \left( K \cdot \sum_{i=1}^{K} f_i^{(l)} \cdot Q_i^{(l)} \right),
  \end{equation}
  \begin{equation}
    f_i^{(l)} = \frac{1}{N} \sum_{v=1}^{N} \mathbb{I}\{i \in \widehat{\mathcal{F}}_v^{(l)}\}, \quad Q_i^{(l)} = \frac{1}{N} \sum_{v=1}^{N} \pi_{v,i}^{(l)}.
  \end{equation}
  where $f_i^{(l)}$ represents the \textit{actual selection frequency} of expert $E_i^{(l)}$, $\mathbb{I}\{\cdot\}$ denotes the indicator function, and $Q_i^{(l)}$ is the average routing probability assigned to $f_i^{(l)}$.
  It is worth noting that $f_i^{(l)}$ is non-differentiable, thus the gradient of $\mathcal{L}_\text{LB}$ is back-propagated solely through the differentiable term $Q_i^{(l)}$, encouraging the router to lower the routing probability scores of overloaded experts.
\end{itemize}

\subsubsection{Total Objective}
The final objective function is a weighted sum of the task loss and the regularization terms:
\begin{equation}
    \mathcal{L}= \mathcal{L}_\text{task} + \lambda_1 \cdot \mathcal{L}_\text{RE} + \lambda_2 \cdot \mathcal{L}_\text{LB},
\end{equation}
where $\lambda_1$ and $\lambda_2$ control the strength of the sparsity and load balancing constraints, respectively.

\section{Experiments}
\label{sec:experiments}
\subsection{Experimental Setup}
\subsubsection{Datasets} We conduct experiments on 13 benchmarks, covering both homophilous and heterophilous scenarios:
(1) \textbf{Homophilous:} Computers, Photo~\cite{mcauley2015image}, Coauthor CS, Coauthor Physics~\cite{shchur2018pitfalls}, Wiki-CS, Facebook~\cite{rozemberczki2021multi}, and Ogbn-arxiv~\cite{hu2020open};
(2) \textbf{Heterophilous:} Actor~\cite{tang2009S}, Chameleon-filtered, Squirrel-filtered, Roman-empire~\cite{critical}, Tolokers, Penn94~\cite{LINKX}.
We adopt a standard 48\%/32\%/20\% split for training, validation, and testing, except for Ogbn-arxiv which follows the official OGB~\cite{hu2020open} public splits.

\subsubsection{Baselines} We compare \model with 19 baselines categorized into four groups: 
(1) \textbf{Vanilla Models:} MLP, GCN~\cite{kipf2017semisupervised}, GraphSAGE~\cite{hamilton2017inductive}, GAT~\cite{velivckovic2017gat};
(2) \textbf{Heterophilic GNNs:} H2GCN~\cite{H2GCN}, GPRGNN~\cite{GPRGNN}, FAGCN~\cite{FAGCN}, ACMGCN~\cite{luan2022revisiting}, FSGNN~\cite{FSGNN}; 
(3) \textbf{Graph Transformers (GTs):} Vanilla GT, ANS-GT~\cite{ASN-GT}, NAGphormer~\cite{NAGphormer}, SGFormer~\cite{SGFormer}, Exphormer~\cite{Exphormer}, Difformer~\cite{DIFFormer}; 
(4) \textbf{Graph MoEs:} GMoE~\cite{GMoE}, DAMoE~\cite{DAMoE}, NodeMoE~\cite{NodeMoE}, Mowst~\cite{Mowst}, Moscat~\cite{Moscat}. 

\begin{table*}[!htb]
  \centering
  \caption{Node classification results: average test accuracy (\%) $\pm$ standard deviation. Boldface letters mark the best performance while underlined letters indicate the second best.}
  \label{tab:main_results}
  \begin{sc}
  \renewcommand\arraystretch{1.2}
  \resizebox{\textwidth}{!}{
  \begin{tabular}{c|c|ccccccc|cccccc} 
  \hline\hline
  \multicolumn{2}{l|}{\diagbox{\textbf{Method}}{\textbf{Dataset}}}        & \textbf{Computers}    & \textbf{Photo}        & \begin{tabular}[c]{@{}c@{}}\textbf{Coauthor}\\\textbf{CS}\end{tabular} & \begin{tabular}[c]{@{}c@{}}\textbf{Coauthor}\\\textbf{Physics}\end{tabular} & \textbf{Wiki-CS}      & \textbf{Facebook}     & \textbf{Ogbn-arxiv}   & \textbf{Actor}         & \begin{tabular}[c]{@{}c@{}}\textbf{Chameleon}\\\textbf{-fix}\end{tabular} & \begin{tabular}[c]{@{}c@{}}\textbf{Squirrel}\\\textbf{-fix}\end{tabular} & \textbf{Tolokers} & \begin{tabular}[c]{@{}c@{}}\textbf{Roman}\\\textbf{-empire}\end{tabular}       & \textbf{Penn94}        \\ 
  \hline
  \multirow{4}{*}{\rotatebox{90}{Vanilla}}       & MLP        & 85.01 $\pm$ 0.84 & 92.00 $\pm$ 0.56 & 94.80 $\pm$ 0.35     & 96.11 $\pm$ 0.14        & 79.57 $\pm$ 0.81 & 76.86 $\pm$ 0.34 & 53.46 $\pm$ 0.35 & 37.14 $\pm$ 1.06  & 33.31 $\pm$ 2.32        & 34.47 $\pm$ 3.09 & 53.18 $\pm$ 6.35 & 65.98 $\pm$ 0.43       & 75.18 $\pm$ 0.35  \\
                                                 & GCN        & 91.17 $\pm$ 0.54 & 94.26 $\pm$ 0.59 & 93.40 $\pm$ 0.45     & 96.37 $\pm$ 0.20        & 83.80 $\pm$ 0.66 & 93.98 $\pm$ 0.34 & 69.71 $\pm$ 0.18 & 30.65 $\pm$ 1.06  & 41.85 $\pm$ 3.22        & 33.89 $\pm$ 2.61 & 70.34 $\pm$ 1.64 & 50.76 $\pm$ 0.46       & 80.45 $\pm$ 0.27  \\
                                                 & GAT        & 91.44 $\pm$ 0.43 & 94.42 $\pm$ 0.61 & 93.20 $\pm$ 0.64     & 96.28 $\pm$ 0.31        & 83.99 $\pm$ 0.73 & 94.03 $\pm$ 0.36 & 70.03 $\pm$ 0.42 & 30.58 $\pm$ 1.18  & 43.31 $\pm$ 3.42        & 36.27 $\pm$ 2.12 & 79.93 $\pm$ 0.77 & 57.34 $\pm$ 1.81       & 78.10 $\pm$ 1.28  \\
                                                 & SAGE       & 90.94 $\pm$ 0.56 & 95.41 $\pm$ 0.45 & 94.17 $\pm$ 0.46     & 96.69 $\pm$ 0.23        & 84.75 $\pm$ 0.64 & 93.72 $\pm$ 0.35 & 69.15 $\pm$ 0.18 & 37.60 $\pm$ 0.95  & 44.94 $\pm$ 3.67        & 36.61 $\pm$ 3.06 & 82.37 $\pm$ 0.64 & 77.77 $\pm$ 0.49       & OOM           \\ 
  \hline
  \multirow{5}{*}{\rotatebox{90}{Hetero}} & H2GCN-2020      & 91.69 $\pm$ 0.33                            & 95.59 $\pm$ 0.48                            & 95.62 $\pm$ 0.27                      & 97.00 $\pm$ 0.16                           & 84.62 $\pm$ 0.66                            & 94.36 $\pm$ 0.32                            & OOM                                     & 37.27 $\pm$ 1.27                            & 43.09 $\pm$ 3.85   & 40.07 $\pm$ 2.73          & 81.34 $\pm$ 1.16     & 79.47 $\pm$ 0.43                                                         & 75.91 $\pm$ 0.44                             \\
                                                 & GPRGNN-2021     & 91.80 $\pm$ 0.55                            & 95.44 $\pm$ 0.33                            & 95.17 $\pm$ 0.34                                         & 96.94 $\pm$ 0.20                                              & 84.84 $\pm$ 0.54                            & 94.84 $\pm$ 0.24                            & 69.95 $\pm$ 0.19                            & 36.89 $\pm$ 0.83                            & 44.27 $\pm$ 5.23                                            & 40.58 $\pm$ 2.00                       & 73.84 $\pm$ 1.40     & 67.72 $\pm$ 0.63                                                         & {\cellcolor[rgb]{0.961,1,0.659}}84.34 $\pm$ 0.29                             \\
                                                 & FAGCN-2021      & 89.54 $\pm$ 0.75                            & 94.44 $\pm$ 0.62                            & 94.93 $\pm$ 0.22                                         & 96.91 $\pm$ 0.27                                              & 84.47 $\pm$ 0.75                            & 91.90 $\pm$ 1.95                            & 66.87 $\pm$ 1.48                            & 37.59 $\pm$ 0.95                            & 45.28 $\pm$ 4.33                                            & 41.05 $\pm$ 2.67                       & 81.38 $\pm$ 1.34     & 75.83 $\pm$ 0.35                                                         & 79.01 $\pm$ 1.09                             \\
                                                 & ACMGCN-2022     & 91.66 $\pm$ 0.78                            & 95.42 $\pm$ 0.39                            & 95.47 $\pm$ 0.33                                         & 97.00 $\pm$ 0.27                           & 85.10 $\pm$ 0.77                            & 94.27 $\pm$ 0.33                            & 69.98 $\pm$ 0.11                            & 36.89 $\pm$ 1.13                            & 43.99 $\pm$ 2.02                                            & 36.58 $\pm$ 2.75          & 83.52 $\pm$ 0.87     & 81.57 $\pm$ 0.35                                                         & 83.01 $\pm$ 0.46                             \\
                                                 & FSGNN-2022      & 91.03 $\pm$ 0.56                            & 95.50 $\pm$ 0.41                            & 95.51 $\pm$ 0.32                                         & 96.98 $\pm$ 0.20                                              & 85.10 $\pm$ 0.58                            & 94.32 $\pm$ 0.32                            & 71.09 $\pm$ 0.21                            & 37.14 $\pm$ 1.06                            & 45.79 $\pm$ 3.31                                            & 38.25 $\pm$ 2.62                       & 83.87 $\pm$ 0.98     & 79.76 $\pm$ 0.41                                                         & 83.87 $\pm$ 0.98                             \\
  \hline

  \multirow{6}{*}{\rotatebox{90}{GT}}    & Vanilla GT & 84.41 $\pm$ 0.72                            & 91.58 $\pm$ 0.73                            & 94.61 $\pm$ 0.30                                         & OOM                                                       & 79.05 $\pm$ 0.97                            & OOM                                     & OOM                                     & 37.08 $\pm$ 1.08                            & 44.27 $\pm$ 3.98                                            & 39.55 $\pm$ 3.10                                         & 72.24 $\pm$ 1.17    & OOM                                                                                    & OOM                                      \\
                                         & ANS-GT-2022     & 90.01 $\pm$ 0.38                            & 94.51 $\pm$ 0.24                            & 93.93 $\pm$ 0.23                                         & 96.28 $\pm$ 0.19                                              & 83.27 $\pm$ 0.49                            & 92.61 $\pm$ 0.16                            & OOM                                     & 37.80 $\pm$ 0.95                            & 40.74 $\pm$ 2.26                                            & 36.65 $\pm$ 0.80                            & 76.91 $\pm$ 0.85    & 80.36 $\pm$ 0.71                                                                       & OOM                                      \\
                                         & NAGFormer-2023  & 90.22 $\pm$ 0.42                            & 94.95 $\pm$ 0.52                            & 94.96 $\pm$ 0.25                                         & 96.43 $\pm$ 0.20                                              & 84.31 $\pm$ 0.72                            & 93.35 $\pm$ 0.28                            & 70.25 $\pm$ 0.13                            & 36.99 $\pm$ 1.39                            & 46.12 $\pm$ 2.25                                            & 38.31 $\pm$ 2.43                        & 66.73 $\pm$ 1.18    & 75.92 $\pm$ 0.69                                                                       & 73.98 $\pm$ 0.53                             \\
                                         & Exphormer-2023  & 91.46 $\pm$ 0.51                            & 95.42 $\pm$ 0.26                            & 95.62 $\pm$ 0.29                      & 96.89 $\pm$ 0.20                                              & 84.57 $\pm$ 0.58                            & 93.88 $\pm$ 0.40                            & 71.59 $\pm$ 0.24                            & 36.83 $\pm$ 1.10                            & 42.58 $\pm$ 3.24                         & 36.19 $\pm$ 3.20                              & 82.26 $\pm$ 0.41    & {\cellcolor[rgb]{0.961,1,0.659}}87.55 $\pm$ 1.13                                       & OOM                                      \\
                                         & Difformer-2023  & 91.52 $\pm$ 0.55                            & 95.41 $\pm$ 0.38                            & 95.49 $\pm$ 0.26                                         & 96.98 $\pm$ 0.22                                              & 83.54 $\pm$ 0.70                            & 94.23 $\pm$ 0.47                            & OOM                                     & 36.73 $\pm$ 1.27                            & 44.44 $\pm$ 3.20                                            & 40.45 $\pm$ 2.51                            & 81.04 $\pm$ 4.16    & 78.97 $\pm$ 0.54                                                                       & OOM                                      \\ 
                                         & SGFormer-2024   & 90.70 $\pm$ 0.59                            & 94.46 $\pm$ 0.49                            & 95.21 $\pm$ 0.20                                         & 96.87 $\pm$ 0.18                                              & 82.67 $\pm$ 0.58                            & 86.66 $\pm$ 0.54                            & 65.84 $\pm$ 0.24                            & 36.59 $\pm$ 0.90                            & 44.27 $\pm$ 3.68                                            & 38.83 $\pm$ 2.19                        & 80.46 $\pm$ 0.91    & 76.41 $\pm$ 0.50                                                                       & 76.65 $\pm$ 0.49                             \\
  \hline

  \multirow{4}{*}{\rotatebox{90}{\scriptsize Graph MoE}}         & GMoE-2023       & 91.37 $\pm$ 0.49                            & 94.51 $\pm$ 0.68                                   & 93.18 $\pm$ 0.58                                         & 96.48 $\pm$ 0.23                                              & 83.65 $\pm$ 0.61                                         & 94.90 $\pm$ 0.25                            & {\cellcolor[rgb]{0.961,1,0.659}}71.66 $\pm$ 0.29         & 33.78 $\pm$ 1.32                            & 46.69 $\pm$ 3.55                                            & 42.24 $\pm$ 2.45                & 85.21 $\pm$ 0.40  & 84.78 $\pm$ 0.76 & 79.03 $\pm$ 0.78                             \\
                                                                 & NodeMoE-2024    & 91.87 $\pm$ 0.33                            & 95.53 $\pm$ 0.41   & OOM                                                      & OOM                                                           & 85.08 $\pm$ 0.62                                         & 94.84 $\pm$ 0.28                            & 70.51 $\pm$ 0.29                                         & 36.28 $\pm$ 1.39                            & 45.67 $\pm$ 4.54                                            & 40.49 $\pm$ 2.01                                                & OOM                                               & 74.31 $\pm$ 0.87 & OOM                                      \\
                                                                 & Mowst-2024      & 92.00 $\pm$ 0.43                            & 95.49 $\pm$ 0.36                                   & 94.51 $\pm$ 0.43                                         & 96.73 $\pm$ 0.18~                                             & 85.13 $\pm$ 0.84         & 95.12 $\pm$ 0.24                            & 71.34 $\pm$ 0.24                                         & 37.68 $\pm$ 0.73                            & 44.16 $\pm$ 3.55                                            & 38.52 $\pm$ 2.07                                                & 79.43 $\pm$ 0.66                                  & 79.56 $\pm$ 0.51 & 78.84 $\pm$ 1.99                             \\ 
                                                                 & DAMoE-2025      & 91.57 $\pm$ 0.64                            & 94.39 $\pm$ 0.53                                   & 93.42 $\pm$ 0.50                                         & 96.42 $\pm$ 0.28                                              & 84.04 $\pm$ 0.80                                         & 94.96 $\pm$ 0.21                            & 71.50 $\pm$ 0.22                                         & 28.76 $\pm$ 1.01                            & 45.51 $\pm$ 2.80                                            & 41.08 $\pm$ 2.08                                                & 51.45 $\pm$ 1.07                                  & 81.92 $\pm$ 0.52 & 78.04 $\pm$ 0.58                             \\& Moscat-2025      & 91.65 $\pm$ 0.59                            & 95.48 $\pm$ 0.47                                   & 94.78 $\pm$ 0.40                                         & {\cellcolor[rgb]{0.961,1,0.659}}97.11 $\pm$ 0.16                                              & 81.40 $\pm$ 0.64                                         & 94.46 $\pm$ 0.53                            & 70.37 $\pm$ 0.24                                         & 35.50 $\pm$ 1.67                            & 43.37 $\pm$ 2.60                                            & 42.25 $\pm$ 2.66                                                & 81.79 $\pm$ 1.07                                  & 77.96 $\pm$ 0.37 & 83.91 $\pm$ 0.63                             \\
  \hline
  \multicolumn{2}{c|}{\textbf{\model (GCN)}}               & {\cellcolor[rgb]{0.663,1,1}}\textbf{92.23 $\pm$ 0.33} & 95.59 $\pm$ 0.52                            & {\cellcolor[rgb]{0.663,1,1}}\textbf{95.82 $\pm$ 0.15}             & {\cellcolor[rgb]{0.663,1,1}}\textbf{97.14 $\pm$ 0.21}                   & 84.81 $\pm$ 0.65                            & 95.36 $\pm$ 0.26         & {\cellcolor[rgb]{0.663,1,1}}\textbf{71.72 $\pm$ 0.25} & {\cellcolor[rgb]{0.961,1,0.659}}38.01 $\pm$ 1.15         & {\cellcolor[rgb]{0.663,1,1}}\textbf{50.39 $\pm$ 4.71}                 & {\cellcolor[rgb]{0.663,1,1}}\textbf{44.11 $\pm$ 3.53}  & {\cellcolor[rgb]{0.663,1,1}}\textbf{85.74 $\pm$ 0.51}     & 84.39 $\pm$ 0.55        & {\cellcolor[rgb]{0.663,1,1}}\textbf{84.55 $\pm$ 0.51}  \\
  \multicolumn{2}{c|}{\textbf{\model (SAGE)}}              & 92.01 $\pm$ 0.38         & {\cellcolor[rgb]{0.961,1,0.659}}95.64 $\pm$ 0.28 & 95.42 $\pm$ 0.17                                        & 96.89 $\pm$ 0.21                                              & {\cellcolor[rgb]{0.961,1,0.659}}85.23 $\pm$ 0.65 & {\cellcolor[rgb]{0.663,1,1}}\textbf{95.47 $\pm$ 0.37} & 71.36 $\pm$ 0.17                            & {\cellcolor[rgb]{0.663,1,1}}\textbf{38.26 $\pm$ 1.06} & {\cellcolor[rgb]{0.961,1,0.659}}49.16 $\pm$ 5.39                         & 40.11 $\pm$ 2.94                        & 84.63 $\pm$ 0.81                                         & {\cellcolor[rgb]{0.663,1,1}}\textbf{87.61 $\pm$ 0.26}                           & 81.02 $\pm$ 0.63                             \\
    \multicolumn{2}{c|}{\textbf{\model (GAT)}}              & {\cellcolor[rgb]{0.961,1,0.659}}92.14 $\pm$ 0.39         & {\cellcolor[rgb]{0.663,1,1}}\textbf{95.79 $\pm$ 0.47} & {\cellcolor[rgb]{0.961,1,0.659}}95.71 $\pm$ 0.25                                        & 97.07 $\pm$ 0.26                                              & {\cellcolor[rgb]{0.663,1,1}}\textbf{85.45 $\pm$ 0.74} & {\cellcolor[rgb]{0.961,1,0.659}}95.44 $\pm$ 0.41 & 71.41 $\pm$ 0.24                            & 37.53 $\pm$ 0.88 & 48.31 $\pm$ 4.74                         &{\cellcolor[rgb]{0.961,1,0.659}} 42.40 $\pm$ 2.38                        & {\cellcolor[rgb]{0.961,1,0.659}}85.26 $\pm$ 0.81                                         & 84.09 $\pm$ 0.34                           & 82.86 $\pm$ 0.47                             \\
  \hline\hline
  \end{tabular}}
\end{sc}
\end{table*}

\subsubsection{Implementation Details}
We report the average test accuracy and standard deviation over 10 runs with different random seeds.
We search hyperparameters for all methods within a unified space: hidden dimension $h=64$, learning rate $\in$ \{0.005, 0.01, 0.05, 0.1\}, dropout $\in$ \{0.1, 0.3, 0.5, 0.7, 0.9\}. The maximum epochs are set to 500 with an early stopping patience of 100. We employ AdamW optimizer and weight decay. 
For \model, the parameter space is: regularization weights $\lambda_1=0.0001$, $\lambda_2\in$ \{1e-3, 1e-1\}, sensitivity coefficient $\gamma\in$\{1, 5, 10\}, maximum number of experts $K\in$\{4, 6\}, maximum number of layers $L\in$\{1, 2\}.  

\begin{figure*}[!htb]
  \centering
  \includegraphics[width=\linewidth]{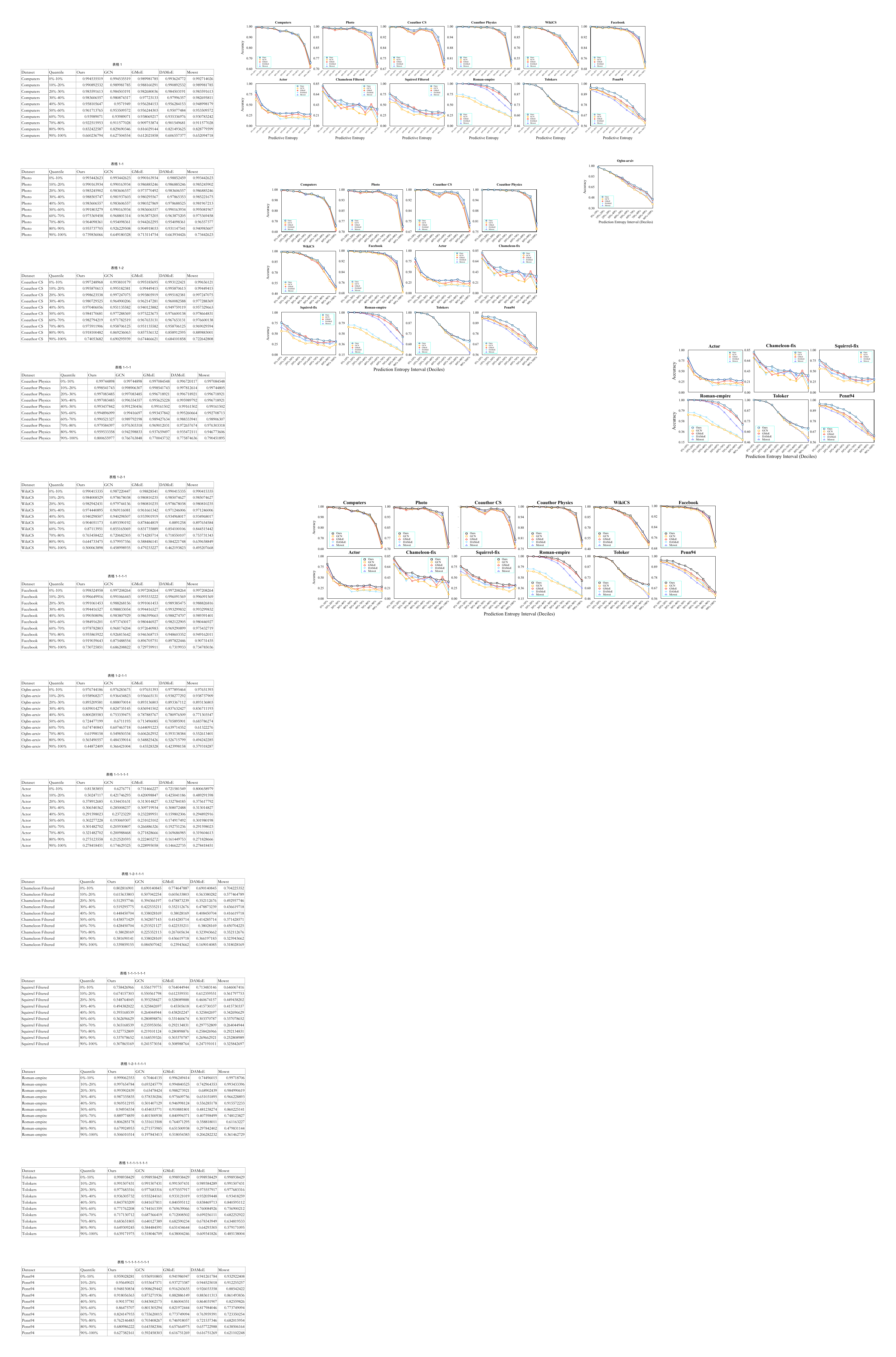}
  \caption{Fine-grained performance comparison across different node difficulty levels. We stratify test nodes into 10 equal-sized intervals (deciles) based on their predictive entropy (calculated by a fixed proxy model). The x-axis represents entropy percentiles (from Easy to Hard), and the y-axis denotes classification accuracy. \model demonstrates superior robustness, particularly in high-entropy (hard) intervals.}
  \label{fig: group-acc}
\end{figure*}

\subsection{Node Classification Performance Analysis}

\subsubsection{Overall Performance Analisys}
As evidenced in Table~\ref{tab:main_results}, \model establishes a new state-of-the-art by outperforming 19 baselines across all 13 datasets. Notably, the performance advantage is most pronounced on heterophilous graphs (\eg \textit{Squirrel}, \textit{Actor}, \textit{Roman-empire}), where \model surpasses the strongest baselines by margins ranging from 0.07\% to 7.92\%.

Crucially, \model demonstrates cross-paradigm superiority, simultaneously transcending the rigid capacity limits of static Graph MoEs, resolving structural ambiguity better than Heterophilic GNNs, and avoiding the optimization bottlenecks of heavy Graph Transformers.
We attribute these substantial gains to the model's sensitivity to structural and semantic ambiguity, which improves its ability to classify hard nodes. This claim is supported by the stratified analysis as follows.

\subsubsection{Fine-grained Performance Analysis across Difficulty Levels}
To provide a deeper insight into how \model handles nodes of varying discriminative difficulty compared to baselines, we conduct a stratified performance analysis. To ensure a rigorous and unbiased comparison, we utilizes a pre-trained, independent proxy model to calculate the predictive entropy for all test nodes (same as motivational experiment settings in Sec.~\ref{sec: obs}). Based on these predictive entropy values, we sort and partition the test nodes into 10 equal-sized intervals (deciles). The intervals range from [0\%-10\%] (lowest entropy, easiest nodes) to [90\%-100\%] (highest entropy, hardest nodes). We then evaluate the average classification accuracy of \model, GCN, and three representative Graph MoE baselines (GMoE, DAMoE, and Mowst) within each interval.

The results across 12 datasets are visualized in Figure~\ref{fig: group-acc}, from which several critical conclusions can be drawn. First, a universal trend observed across all datasets and models is that classification accuracy exhibits a strictly monotonic decrease as predictive entropy increases. This empirical evidence strongly validates our core hypothesis: predictive entropy serves as a reliable and effective proxy for node discriminative difficulty.
Furthermore, while \model maintains comparable performance with baselines in low-entropy intervals ($0\%-30\%$), the performance gap between our method and the baselines becomes significantly more pronounced across the medium-to-high entropy intervals ($40\%-100\%$), indicating that \model consistently achieves higher classification accuracy on these challenging nodes.
This phenomenon highlights the efficacy of our difficulty-aware dynamic routing. For easy nodes with low entropy, \model behaves similarly to baselines and maintains high efficiency. However, for hard nodes with high entropy, the model effectively captures the difficulty via the entropy signal and adaptively scales the computational budget. By activating a denser expert ensemble to reason about these ambiguous samples, this detect and scale capability enables \model to resolve complex patterns that static architectures fail to capture.

\subsection{Ablation Analysis on Routing and Regularization}
\label{sec:ablation}
\begin{figure*}[!htb]
  \centering
  \includegraphics[width=\linewidth]{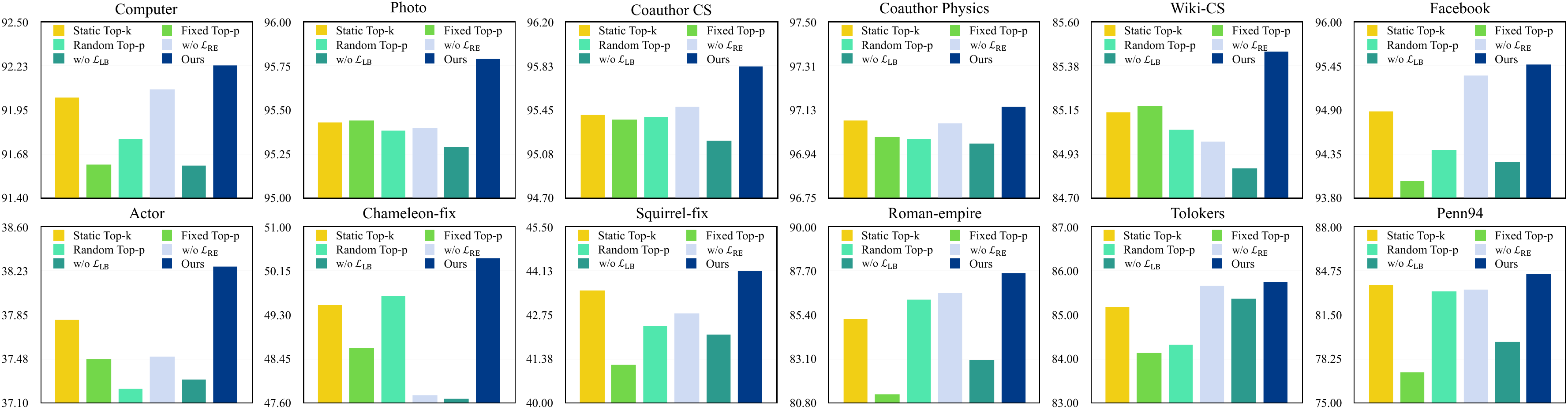} 
  \caption{Ablation analysis on \model.}
  \label{fig:ablation}
\end{figure*}

To assess the contribution of components in \model, we conduct a comprehensive ablation study by comparing \model with four variants:
(1) Static top-$k$, which utilizes a fixed expert budget (searching $k \in \{1,2, 3,4,5,6\}$); 
(2) Fixed top-$p$, which employs a global constant threshold $p$ for all nodes (searching $p \in \{0.3, 0.5, 0.7\}$); 
(3) Random top-$p$, which replaces predictive entropy guidance with random sampling; 
(4) w/o $\mathcal{L}_\text{RE}$, which excludes the routing entropy regularization;
and (5) w/o $\mathcal{L}_\text{LB}$, which removes the load balancing loss.
Results are visualized in Figure~\ref{fig:ablation}.

\begin{figure*}[!htb]
  \centering
  \includegraphics[width=\linewidth]{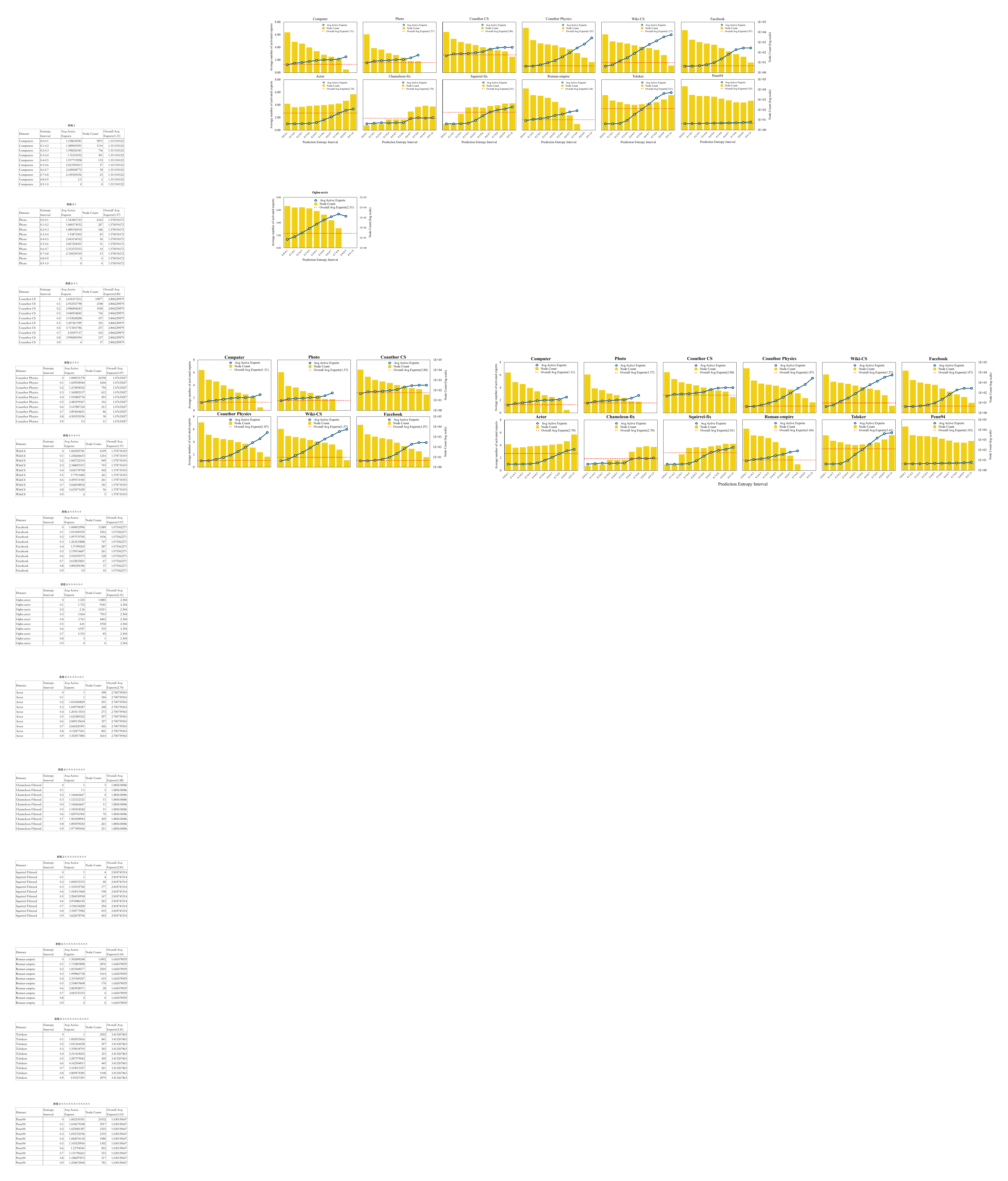} 
  \caption{Node density distribution and average number of activated experts across varying levels of discriminative difficulty.}
  \label{fig: expert_num_vs_difficulty}
\end{figure*}

\subsubsection{Impact of Dynamic Routing} 
We consider the ``Static top-$k$'' and ``Fixed top-$p$'' variants as static routing strategies because the former presumes a uniform computational budget while the latter assumes globally consistent discriminative difficulty. Results indicate that both variants suffer from significant performance degradation with average drops of 1.46\% and 5.00\% on heterophilous graphs. This confirms that rigid architectural priors fail to match the varying difficulty of nodes, leading to a expert resource mismatch, \ie over-computing for easy nodes and under-fitting for hard ones. By breaking these rigid constraints, \model dynamically aligns the computational budget with real-time predictive entropy, effectively resolving this bottleneck. 

\subsubsection{Effectiveness of Difficulty Awareness} 
The ``Random Top-$p$'' variant exhibits an average performance drop of 1.25\% across all datasets. This result refutes the hypothesis that performance gains arise solely from ensemble effects caused by varying expert counts and validates predictive entropy as an effective proxy for discriminative difficulty. Guided by predictive entropy, \model allocates additional computational resources only to hard nodes near decision boundaries, ensuring that extra computation serves to resolve specific decision ambiguities rather than being applied arbitrarily. 

\subsubsection{Necessity of Regularization} 
Both regularization terms are critical for routing stability. Specifically, removing $\mathcal{L}_\text{LB}$ causes a 2.28\% decline due to expert collapse, while removing $\mathcal{L}_\text{RE}$ leads to a 1.13\% drop by encouraging flat routing distributions that introduce noise. Together, these constraints ensure a robust routing that remains globally balanced and locally decisive.

\subsection{Interpretability Analysis of the Routing Mechanism}\label{sec: exp-routing}

To verify whether \model effectively achieves on-demand computation based on discriminative difficulty, we conduct a joint analysis combining macroscopic statistical patterns and microscopic representation distributions.

\subsubsection{Macroscopic Statistical Analysis} 
First, we stratify test nodes into distinct difficulty levels based on their predictive entropy and analyze the node density alongside the average expert activation count. As shown in Figure~\ref{fig: expert_num_vs_difficulty}, \textit{the results reveal \textbf{a significant positive correlation between the computational budget and node discriminative difficulty}, where the average number of active experts exhibits a strict monotonic upward trend as predictive entropy increases.} This confirms that the difficulty-aware routing successfully establishes a dynamic mapping from node-wise difficulty to computational overhead.
Moreover, this adaptive routing mechanism proves robust across diverse graph topologies. Even in heterophilous scenarios characterized by a skew toward high-entropy intervals, \model consistently maintains minimal expert activation for easy nodes to reduce redundancy, while selectively scaling up the budget for hard nodes to ensure sufficient reasoning capacity.

\begin{figure*}[!htbp]
  \centering
  \includegraphics[width=\linewidth]{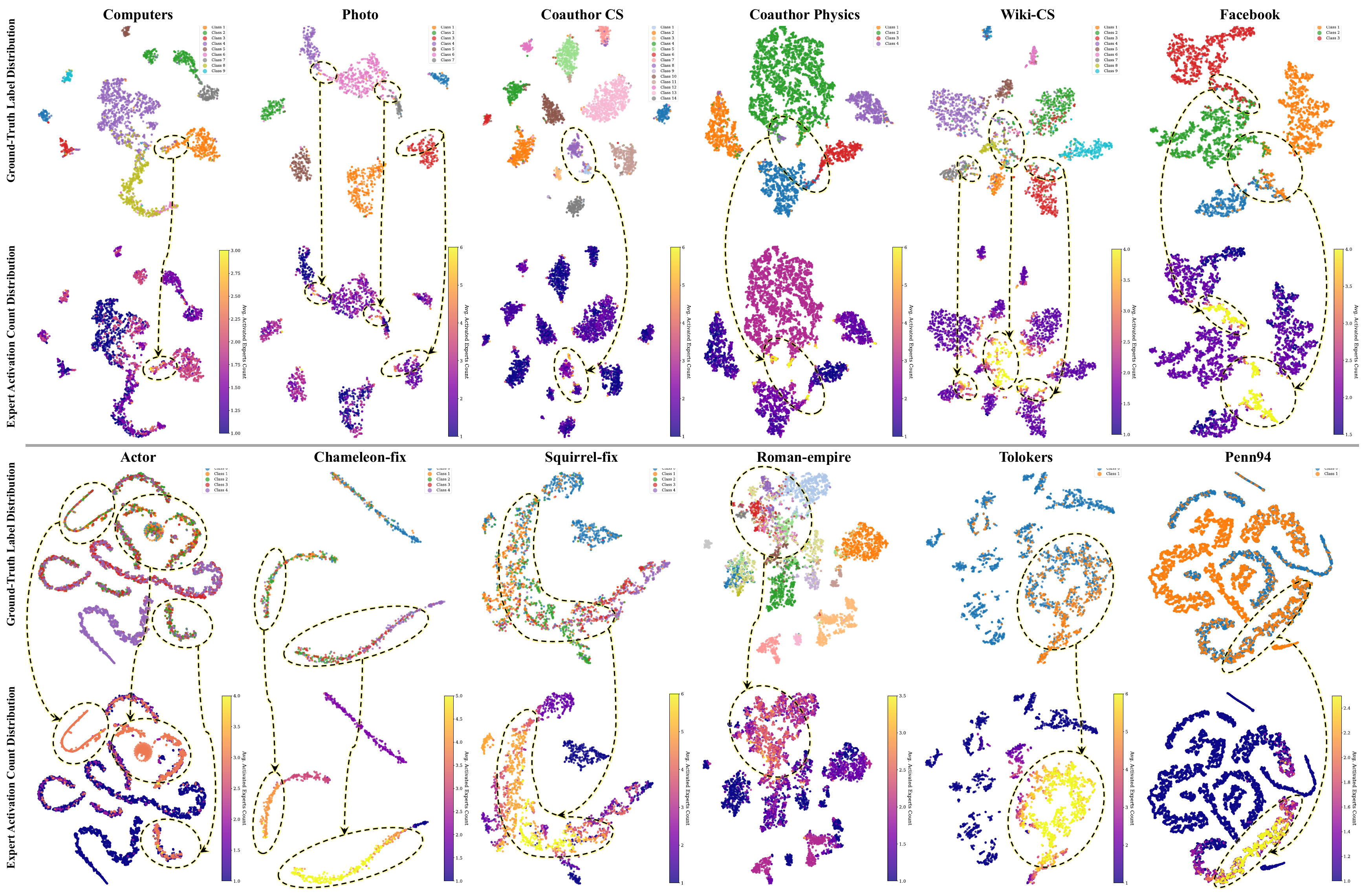}
  \caption{Visualization of learned node representations via t-SNE across 12 datasets. For each dataset, nodes are colored by Ground-Truth Labels (Top) and Activated Expert Counts (Bottom). The results highlight a clear boundary-focused resource allocation pattern.}
  \label{fig: tnse}
\end{figure*}

\subsubsection{Microscopic Visualization Analysis}
To provide microscopic evidence, we present a comprehensive visualization of the learned node representations and the corresponding expert activation patterns across all 12 datasets. Figure~\ref{fig: tnse} presents the t-SNE projections of the node embeddings in the final layer, where each node is colored according to its ground-truth label (Top) and its adaptively allocated expert count (Bottom).

We observe a spatial correspondence between confusion regions in the label distribution space (Top) and high activation regions in the expert count distribution space (Bottom). For homophilous datasets such as \textit{Computers}, \textit{Photo}, and \textit{CS}, the label projections show clearly separated clusters. In these scenarios, \model demonstrates a clear ``core-periphery'' resource allocation strategy. Specifically, cluster centroids where nodes of the same class gather tightly correspond to dark blue areas ($k=1$) in the expert plots. This indicates that \model correctly identifies these clear samples and processes them with minimal computational overhead. Conversely, inter class margins where different clusters meet align perfectly with warmer colors ($k \gg  1$). This visual alignment confirms that the router perceives the semantic ambiguity at these decision boundaries and concentrates resources there to resolve conflicting signals.

In the case of heterophilous graphs, the topology becomes more complex and appears as entangled areas in the label distribution space where class colors mix without clear separation. Accordingly, the expert activation maps shift from localized boundary highlighting to a widespread distribution of warmer colors. This represents a logical generalization because pervasive structural noise turns the majority of the graph into a difficult boundary regions. \model correctly perceives this widespread ambiguity and triggers a global expansion of expert resource to maintain robustness. In summary, \model acts as a reliable detector of semantic ambiguity regardless of whether the boundaries are localized or pervasive. It dynamically scales its reasoning capacity in alignment with the visual difficulty of the data manifold.

\begin{figure*}[!htbp]
  \centering
  \includegraphics[width=\textwidth]{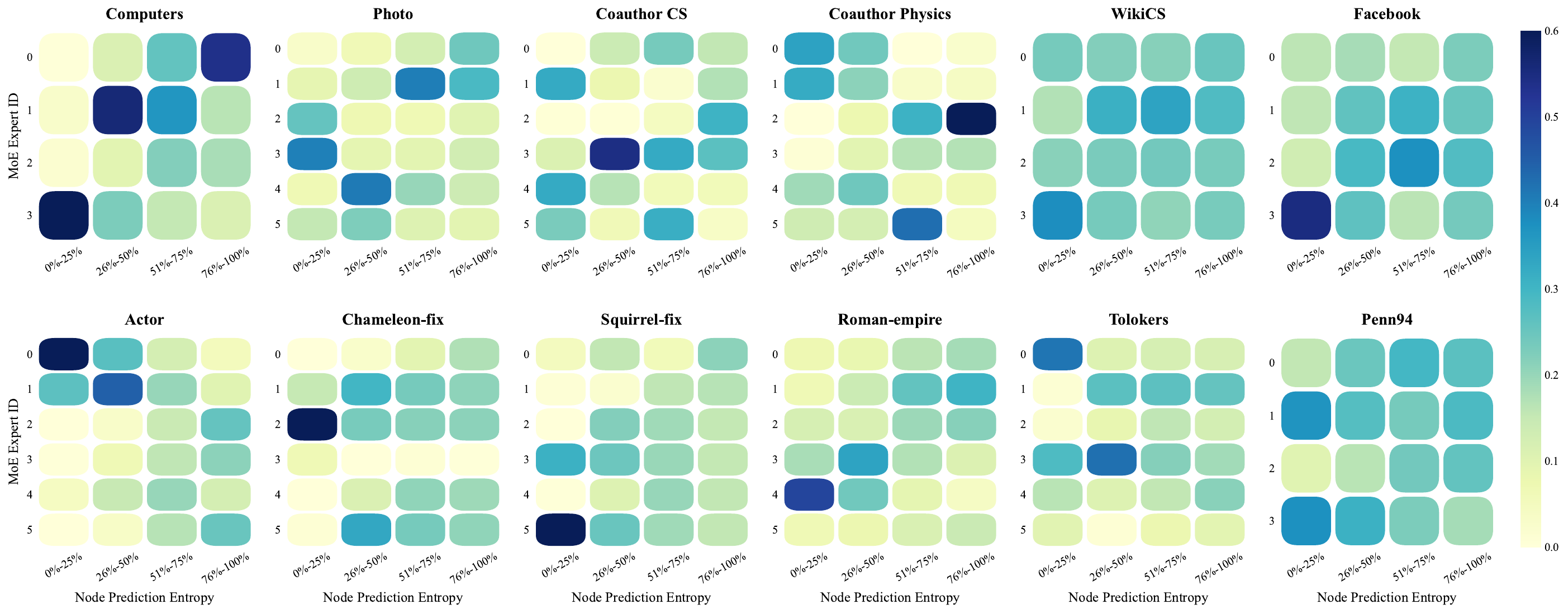} 
  \caption{Visualization of expert activation patterns across 12 datasets. The heatmaps illustrate the average routing weights assigned to different experts (Y-axis) across \textbf{four quartiles of node predictive entropy} (X-axis, ranging from low to high). Brighter colors indicate higher activation weights.}
  \label{fig: heatmap-4}
\end{figure*}

\begin{figure}[!htbp]
  \centering
  \includegraphics[width=\linewidth]{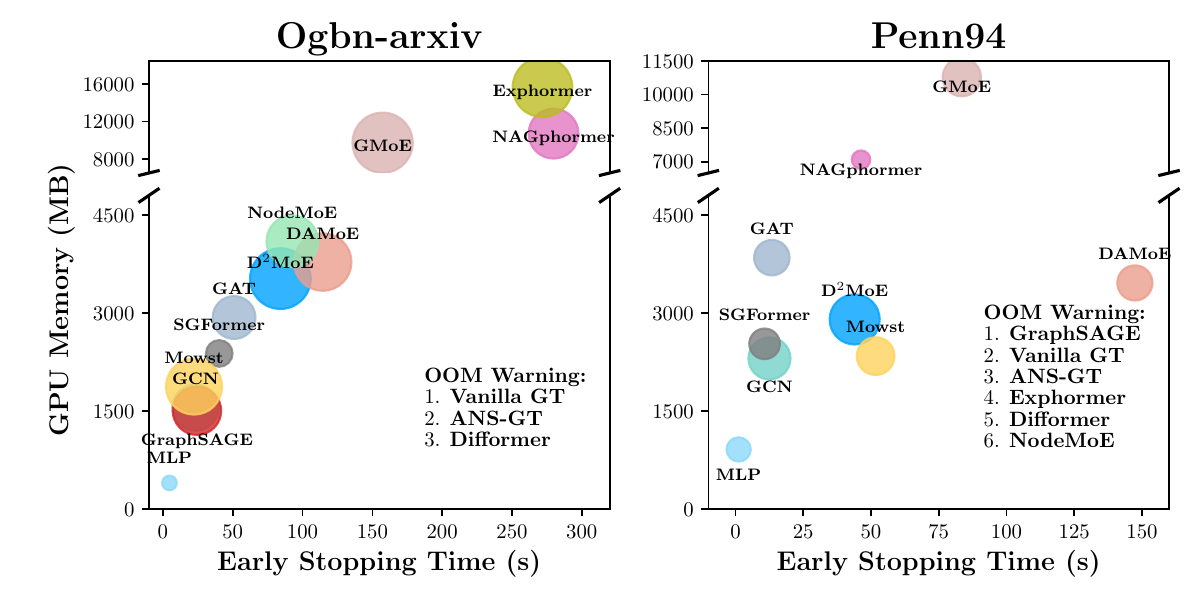}
  \caption{Efficiency analysis on large-scale graphs.}
  \label{fig: efficiency}
\end{figure}

\subsection{Analysis of Expert Activation Patterns}
To investigate the internal routing dynamics within the expert network, we visualize the average routing weights of different experts (Y-axis) across quartile intervals of predictive entropy (X-axis, from low/easy to high/hard), as shown in Figure~\ref{fig: heatmap-4}.

First, we observe distinct functional specialization among experts without explicit supervision. In most datasets (\eg \textit{Computers}, \textit{Actor}, \textit{Squirrel}), specific ``generalist'' experts maintain high activation weights in the lowest entropy intervals ($0\%\sim25\%$). These experts specialize in handling the majority of easy nodes, effectively extracting standard features with high confidence. In contrast, a complementary set of ``specialist'' experts remains inactive in low-entropy regions but becomes highly active in high-entropy intervals ($75\%\sim100\%$). These specialists are activated to resolve complex ambiguities at decision boundaries or in heterophilous neighborhoods. This division of labor effectively mitigates the mode collapse issue often seen in traditional MoEs, where a single expert tends to dominate the processing of all data.

Second, the morphology of the weight distribution evolves distinctly as node difficulty increases, reflecting the model's dynamic expansion of computational capacity. In low-entropy regions, the weight distribution is remarkably sparse and peaky, often concentrated on just one or two dominant generalists. This confirms that for easy nodes, the model confidently relies on a minimal computational path. As discriminative difficulty rises, the weight distribution becomes significantly smoother and more dispersed, spanning a broader range of experts. This transition from ``sparse concentration" to ``dense collaboration" demonstrates that the router automatically expands the expert ensemble size to aggregate diverse perspectives for hard samples. In summary, whether in homophilous graphs or challenging heterophilous scenarios, \model autonomously aligns both the \textbf{role} (Who to compute) and the \textbf{quantity} (How much to compute) of experts with the intrinsic discriminative difficulty of each node.

\subsection{Efficiency Analysis on Large-scale Graphs}
To evaluate the efficiency of \model on large-scale graphs, we conduct a joint analysis of convergence speed, memory consumption, and  accuracy on \textit{Penn94} and \textit{Ogbn-arxiv}. As shown in Figure~\ref{fig: efficiency}, \model successfully achieves an effective balance between computational efficiency and model performance. Although lightweight MoEs like Mowst achieve higher computational efficiency through a minimalist architecture of only two experts, this also limits the model's expressive power, resulting in classification accuracy far inferior to other graph MoEs. Furthermore, in comparison to GTs and other Graph MoEs that frequently encounter Out-Of-Memory (OOM) issues as evidenced in Table~\ref{tab:main_results}, our method significantly reduces peak memory consumption and accelerates convergence while maintaining SOTA accuracy. This stems from the \textit{sparse pruning effect} where \model minimizes the computational budget for the vast majority of easy nodes by often activating only a single expert. Such significant reduction in computational redundancy preserves memory to support deep reasoning for hard nodes, making \model a highly scalable solution for massive graph data.

\section{Conclusion}
In this paper, we presented \model, a difficulty-aware dynamic routing framework that transcends the rigid capacity constraints of static Graph MoEs. By successfully coupling predictive entropy with dynamic expert budget allocation, \model demonstrates that optimal performance arises not merely from increasing model capacity, but from precisely aligning ``thinking effort'' with instance-wise ambiguity. While our framework establishes a solid foundation, we acknowledge technical boundaries regarding potential entropy miscalibration under extreme label noise and the storage overhead of maintaining historical predictions. Accordingly, our future research will focus on addressing these constraints by integrating noise-robust uncertainty metrics, developing lightweight history-free estimators for billion-scale scalability, and extending this adaptive paradigm to time-evolving graph streams.

\appendices
\renewcommand{\thesubsection}{\thesection-\arabic{subsection}}
\renewcommand{\thesubsectiondis}{\thesection-\arabic{subsection}}
\renewcommand{\thesubsubsection}{\thesubsection-\arabic{subsubsection}}
\renewcommand{\thesubsubsectiondis}{\thesubsection-\arabic{subsubsection}}



\bibliographystyle{IEEEtran}
\bibliography{Reference,IEEEabrv}


\vspace{-40pt}

\begin{IEEEbiography}[{\includegraphics[width=1in,height=1.25in,clip,keepaspectratio]{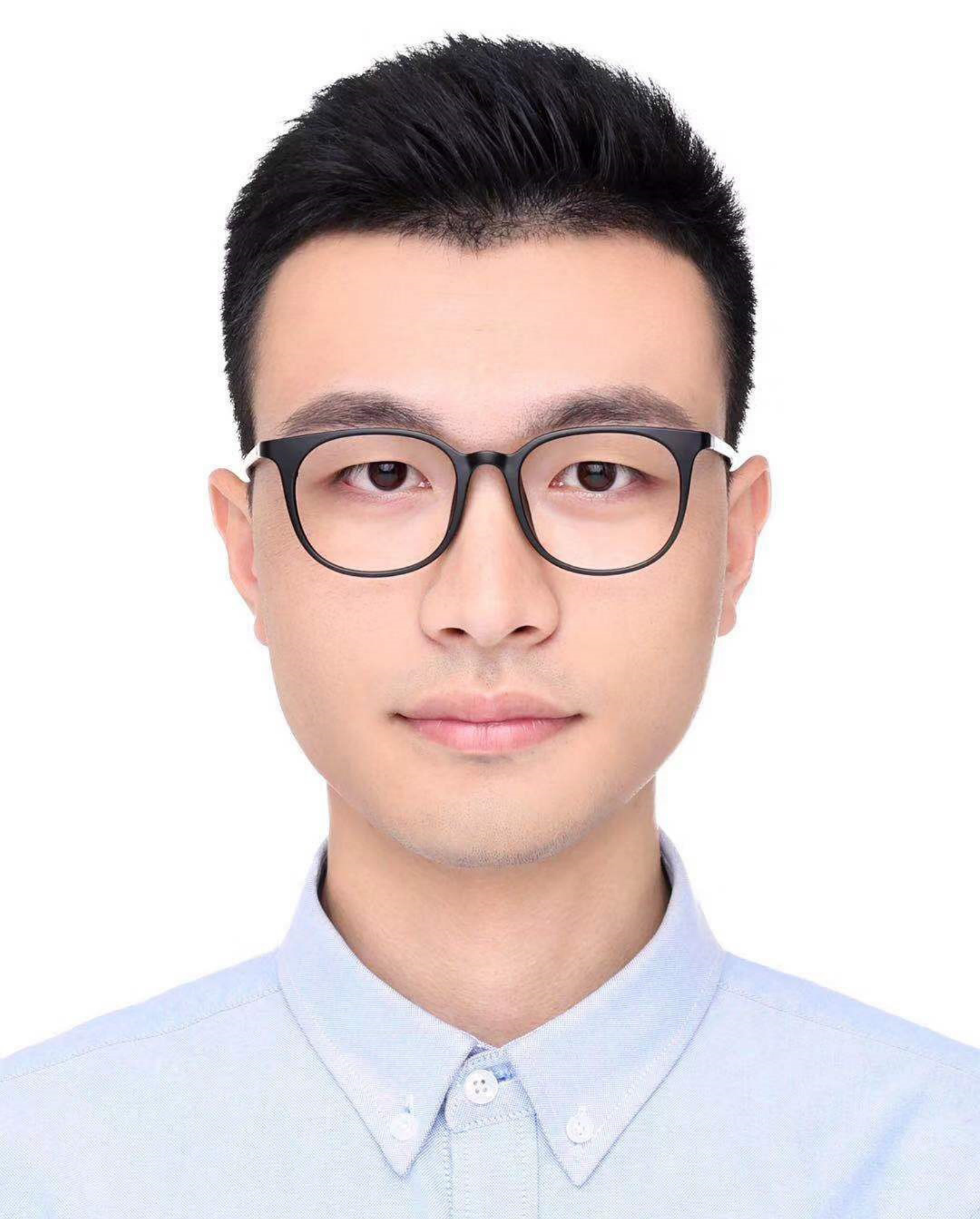}}]{Jiajun Zhou}
	received the Ph.D degree in control theory and engineering from Zhejiang University of Technology, Hangzhou, China, in 2023. He is currently an Associate Research Fellow with the Institute of Cyberspace Security, Zhejiang University of Technology. His current research interests include graph data mining, cyberspace security and AI security.
\end{IEEEbiography}
\vspace{-40pt}

\begin{IEEEbiography}[{\includegraphics[width=1in,height=1.25in,clip,keepaspectratio]{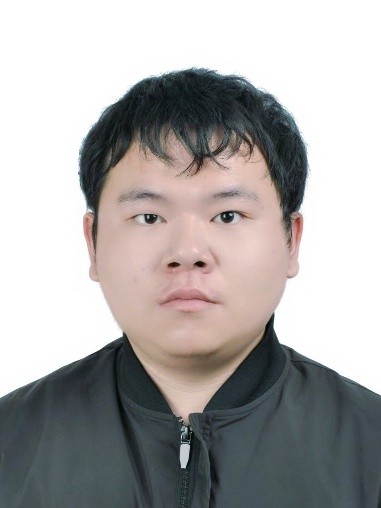}}]{Yadong Li}
	received a BS degree from Tianjin University of Technology, Tianjin, China, in 2024. He is currently pursuing a Master's degree in Control Science and Engineering at Zhejiang University of Technology. His current research interests include graph neural networks and large language models.
\end{IEEEbiography}
\vspace{-40pt}

\begin{IEEEbiography}[{\includegraphics[width=1in,height=1.25in,clip,keepaspectratio]{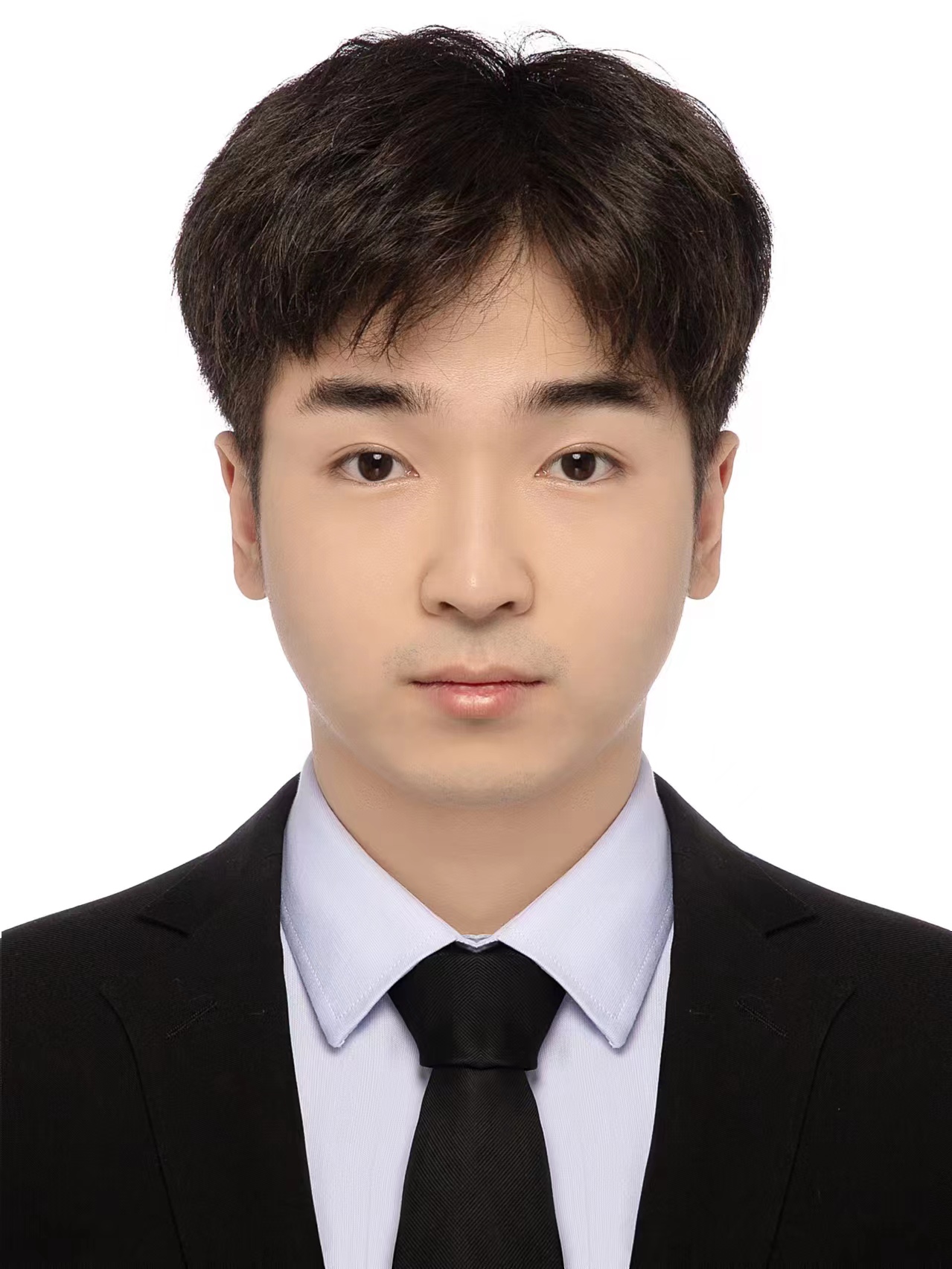}}]{Xuanze Chen}
	received the BS degree from Wenzhou University, Wenzhou, Zhejiang, China, in 2023. He is currently pursuing a Master's degree at the Institute of Cyberspace Security, Zhejiang University of Technology. His current research interests include graph neural networks and Graph Transformers.
\end{IEEEbiography}
\vspace{-40pt}

\begin{IEEEbiography}[{\includegraphics[width=1in,height=1.25in,clip,keepaspectratio]{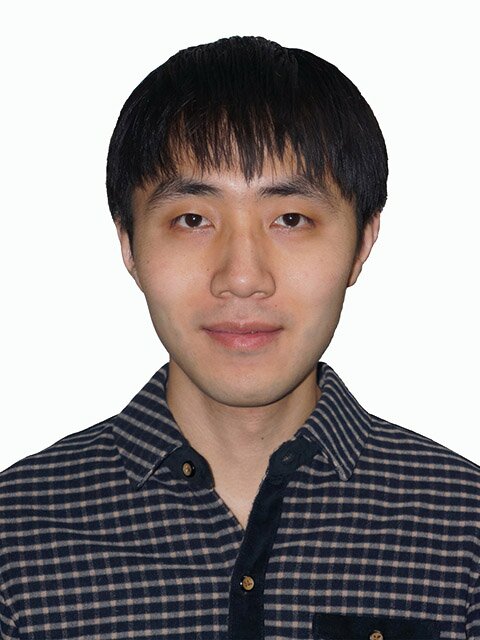}}]{Chen Ma}
	received the M.S. degree from the Beijing University of Posts and Telecommunications, Beijing, China, in 2015, and the Ph.D. degree from Tsinghua University, Beijing, in 2022. From 2009 to 2016, he was a Software Engineer with internet companies, including Sina Ltd., Watford, U.K. Since 2023, he has been with the Institute of Cyberspace Security, Zhejiang University of Technology, Hangzhou, China, where he is currently a Distinguished Research Fellow. His current research interests include adversarial attack and its defense technology, action unit detection, and deep learning interpretability.
\end{IEEEbiography}
\vspace{-40pt}

\begin{IEEEbiography}[{\includegraphics[width=1in,height=1.25in, clip,keepaspectratio]{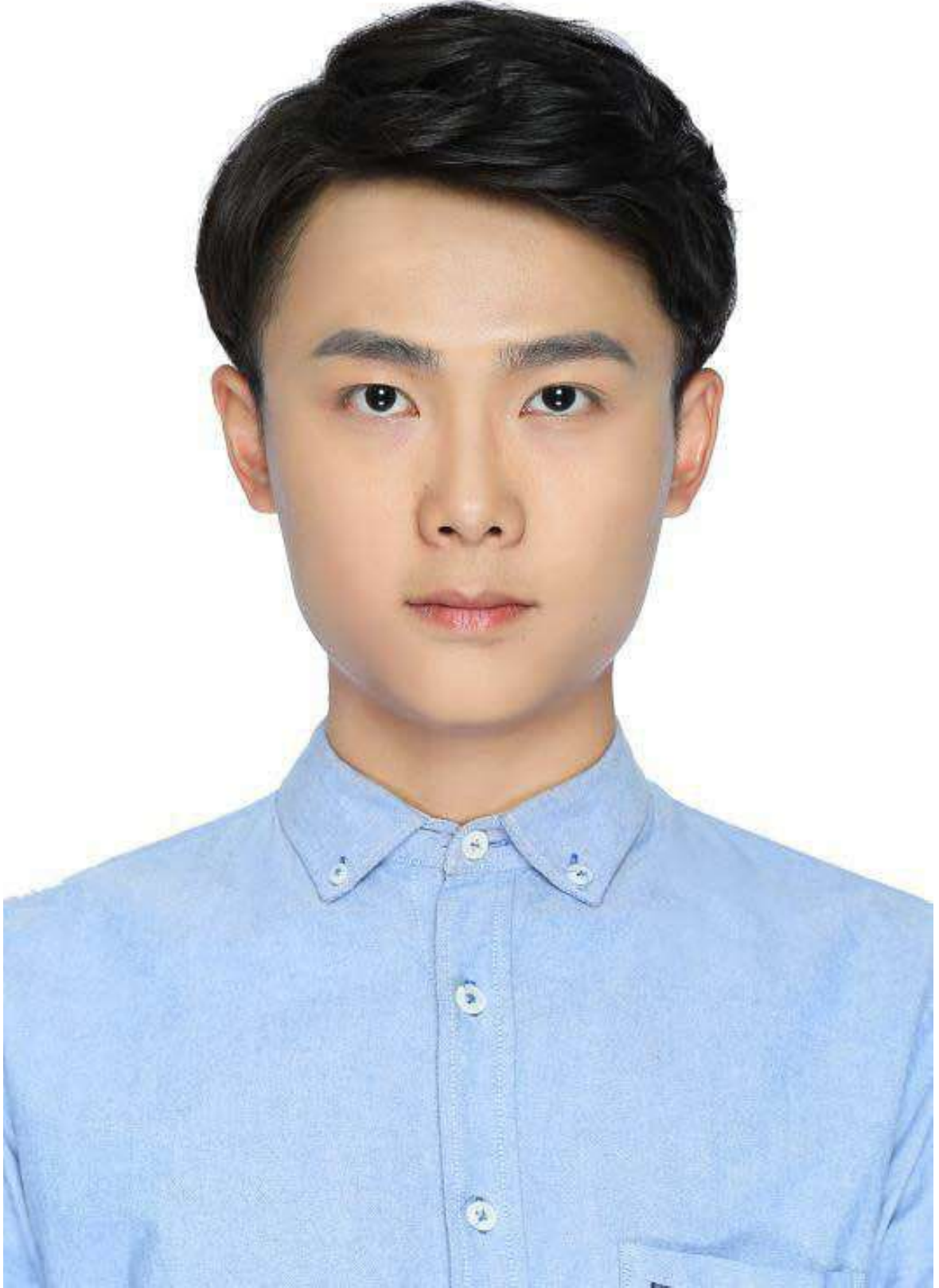}}]{Chuang Zhao} is currently pursuing the Ph.D. degree with Hong Kong University of Science and Technology. He received his master degree from Tianjin University, Tianjin, China. His research interest includes AI for science, recommendation, and data mining. He has published  papers in journals and conference proceedings, including SIGKDD, AAAI, WWW, CIKM, TKDE, TOIS, TEVC, and TNNLS.
\end{IEEEbiography}
\vspace{-40pt}

\begin{IEEEbiography}[{\includegraphics[width=1in,height=1.25in,clip,keepaspectratio]{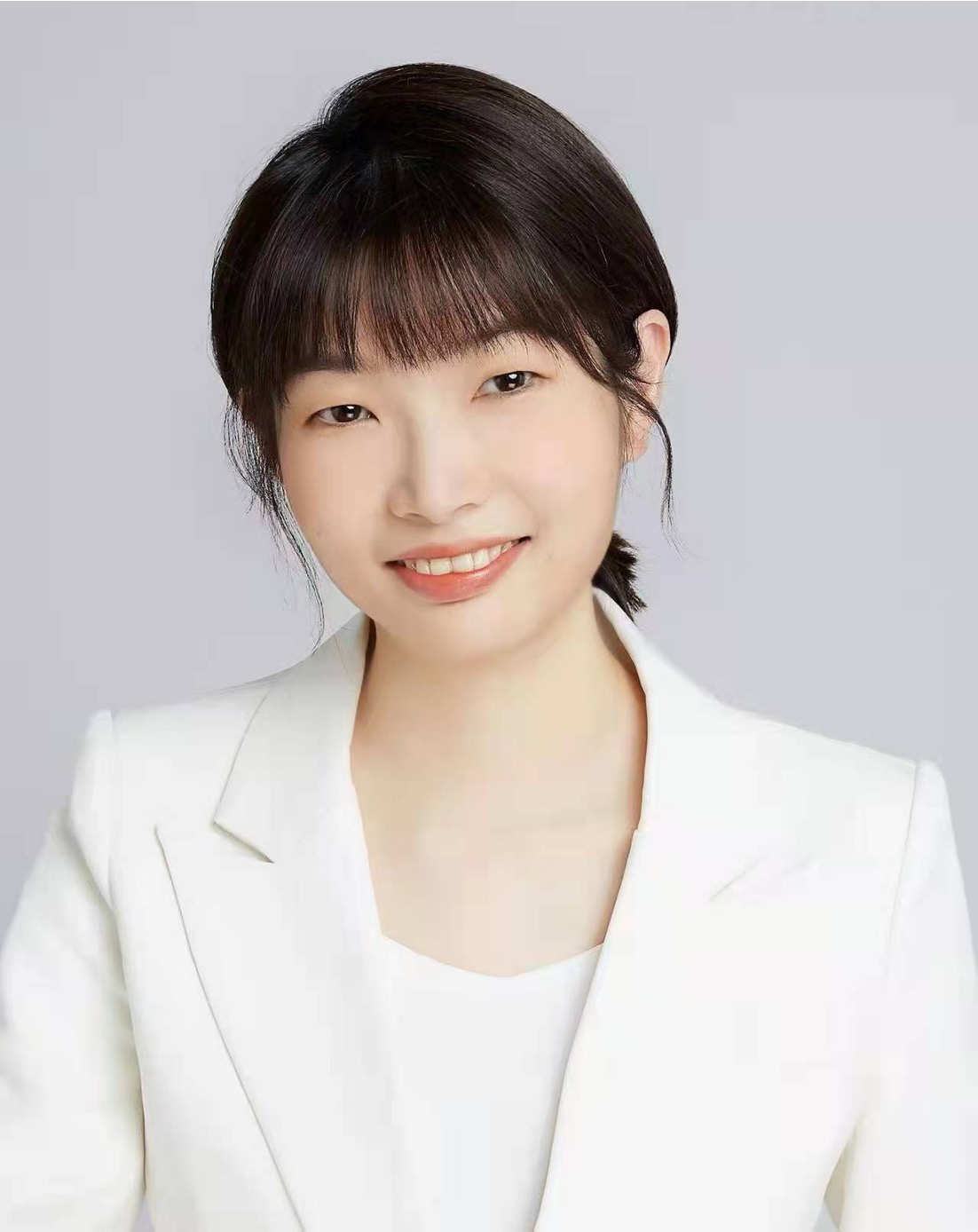}}]{Shanqing Yu}
	received the M.S. degree from the School of Computer Engineering and Science, Shanghai University, China, in 2008 and received the M.S. degree from the Graduate School of Information, Production and Systems, Waseda University, Japan, in 2008, and the Ph.D. degree, in 2011, respectively. She is currently a Lecturer at the Institute of Cyberspace Security and the College of Information Engineering, Zhejiang University of Technology, Hangzhou, China. Her research interests cover intelligent computation and data mining.
\end{IEEEbiography}
\vspace{-35pt}

\begin{IEEEbiography}[{\includegraphics[width=1in,height=1.25in,clip,keepaspectratio]{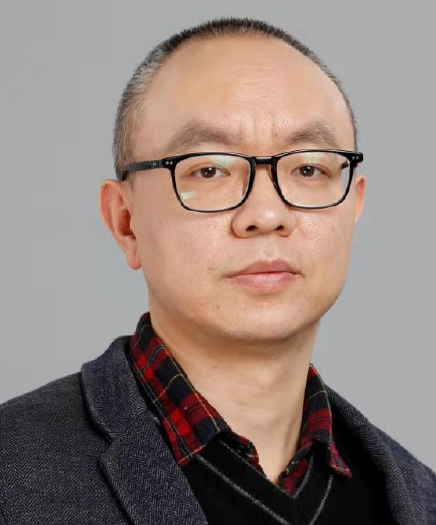}}]{Qi Xuan}(M'18) 
  received the BS and PhD degrees in control theory and engineering from Zhejiang University, Hangzhou, China, in 2003 and 2008, respectively. He was a Post-Doctoral Researcher with the Department of Information Science and Electronic Engineering, Zhejiang University, from 2008 to 2010, respectively, and a Research Assistant with the Department of Electronic Engineering, City University of Hong Kong, Hong Kong, in 2010 and 2017. From 2012 to 2014, he was a Post-Doctoral Fellow with the Department of Computer Science, University of California at Davis, CA, USA. He is a senior member of the IEEE and is currently a Professor with the Institute of Cyberspace Security, College of Information Engineering, Zhejiang University of Technology, Hangzhou, China. His current research interests include network science, graph data mining, cyberspace security, machine learning, and computer vision.
\end{IEEEbiography}


\newpage


\appendices
\renewcommand{\thesubsection}{\thesection-\arabic{subsection}}
\renewcommand{\thesubsectiondis}{\thesection-\arabic{subsection}}
\renewcommand{\thesubsubsection}{\thesubsection-\arabic{subsubsection}}
\renewcommand{\thesubsubsectiondis}{\thesubsection-\arabic{subsubsection}}



\section{Proof of the Generalized Scaling Law}
\label{app:detailed_proof}

Here we provide a comprehensive derivation of \textbf{Theorem 1}. We analyze the generalization error of a sparse Mixture-of-Experts (MoE) system under realistic conditions: weighted expert aggregation and non-zero inter-expert correlation.

\subsection{Generalization Error Decomposition}
Consider a node $v$ with a ground-truth label $y$. Let $X_i$ be the prediction of the $i$-th expert for node $v$, where experts are indexed $i=1, 2, \dots, N$. The dynamic router selects $k$ experts and assigns normalized weights $\pi_i \in [0, 1]$ such that $\sum_{i=1}^k \pi_i = 1$. The ensemble prediction is denoted by $Y_k = \sum_{i=1}^k \pi_i X_i$. The expected Mean Squared Error (MSE) for node $v$ can be decomposed into bias and variance components:
\begin{equation}
    \mathcal{L}(v, k) = \mathbb{E}[(Y_k - y)^2] = \text{Bias}^2(Y_k) + \text{Var}(Y_k) + \epsilon,
\end{equation}
where $\epsilon$ represents the irreducible noise.

\subsection{Derivation of Weighted Bias Accumulation}
The bias of the $i$-th expert is defined as $b_i = \mathbb{E}[X_i] - y$. Without loss of generality, we assume $b_i\ge0$ to analyze the monotonic accumulation of bias, as perfectly canceling opposite biases is overly optimistic in practice. Since \model sorts experts based on their predictive confidence for node $v$, we naturally have $0 \le b_1 \le b_2 \le \dots \le b_N$. The bias of the weighted ensemble is:    
\begin{equation}
    \text{Bias}(Y_k) = \mathbb{E}\left[ \sum_{i=1}^k \pi_i X_i \right] - y = \sum_{i=1}^k \pi_i b_i.
\end{equation}
When the expert budget expands from $k$ to $k+1$, a new weight distribution $\{\pi'_1, \dots, \pi'_{k+1}\}$ is generated. To satisfy the normalization constraint $\sum_{i=1}^{k+1} \pi'_i = 1$, the router must re-allocate probability mass from the initial $k$ experts to the $(k+1)$-th expert. Because $b_{k+1} \ge b_i$ for all $i \in \{1, \dots, k\}$, this re-allocation siphons weight away from more accurate experts and injects a higher-bias component into the average. 

While the Softmax-based weighting scheme in \model assigns decaying weights to lower-ranked experts, thereby mitigating the linear bias growth seen in uniform ensembles, the zero-sum nature of the weights ensures that the ensemble bias remains a strictly monotonic increasing function of $k$. We model this accumulation effect as a generalized power function:
\begin{equation}
    \text{Bias}^2(k) \approx \beta \cdot k^\mu, \quad (\beta > 0, \mu > 0).
\end{equation}

\subsection{Derivation of Correlated Weighted Variance}
We now derive the variance for the weighted ensemble of correlated variables. Let each expert have a prediction variance $\text{Var}(X_i) = \sigma^2$, and assume a constant correlation coefficient $\rho \in (0, 1)$ between any two distinct experts ($i \neq j$), such that $\text{Cov}(X_i, X_j) = \rho \sigma^2$. The variance of the ensemble $Y_k$ is:
\begin{equation}
    \text{Var}(Y_k) = \text{Var}\left( \sum_{i=1}^k \pi_i X_i \right) = \sum_{i=1}^k \pi_i^2 \text{Var}(X_i) + \sum_{i \neq j} \pi_i \pi_j \text{Cov}(X_i, X_j).
\end{equation}
Substituting the correlation term, we obtain:
\begin{equation}
    \text{Var}(Y_k) = \sigma^2 \sum_{i=1}^k \pi_i^2 + \rho \sigma^2 \sum_{i \neq j} \pi_i \pi_j.
\end{equation}
Utilizing the normalization identity $(\sum_{i=1}^k \pi_i)^2 = \sum \pi_i^2 + \sum_{i \neq j} \pi_i \pi_j = 1$, we can substitute the double summation term with $1 - \sum \pi_i^2$:
\begin{equation}
    \text{Var}(Y_k) = \sigma^2 \sum_{i=1}^k \pi_i^2 + \rho \sigma^2 (1 - \sum_{i=1}^k \pi_i^2) = \rho \sigma^2 + (1 - \rho) \sigma^2 \sum_{i=1}^k \pi_i^2.
\end{equation}

In statistical theory, the term $1 / \sum \pi_i^2$ represents the effective ensemble size. For non-uniform weights, the decay rate of $\sum \pi_i^2$ is generalized as $1 / k^\varphi $, where $\varphi  \le 1$. Given that the intrinsic variance $\sigma^2$ is proportional to the predictive uncertainty $\mathcal{U}_v$ (i.e., $\sigma^2 \propto \mathcal{U}_v$), we arrive at the final variance model:
\begin{equation}
    \text{Var}(k) \approx \rho \alpha \mathcal{U}_v + \frac{1 - \rho}{k^\varphi } \alpha \mathcal{U}_v, \quad (\alpha > 0, \varphi  > 0).
\end{equation}
The first term $\rho \alpha \mathcal{U}_v$ represents the \textit{irreducible variance floor} due to inter-expert correlation, while the second term represents the \textit{reducible variance} that diminishes as $k$ increases.

\subsection{Optimization and Scaling Law}
Combining the bias and variance components, the total generalization error is:
\begin{equation}
    \mathcal{L}(v, k) = \beta k^\mu + \rho \alpha \mathcal{U}_v + \frac{(1 - \rho)\alpha \mathcal{U}_v}{k^\varphi } + \epsilon.
\end{equation}
To find the optimal expert budget $k^*$ that minimizes the error, we take the derivative with respect to $k$ and set it to zero:
\begin{equation}
    \frac{\partial \mathcal{L}}{\partial k} = \mu \beta k^{\mu-1} - \varphi (1 - \rho)\alpha \mathcal{U}_v k^{-\varphi -1} = 0.
\end{equation}
Rearranging the terms:
\begin{equation}
    \mu \beta k^{\mu-1} = \frac{\varphi (1 - \rho) \alpha \mathcal{U}_v}{k^{\varphi+1}} \implies k^{\mu+\varphi} = \frac{\varphi (1 - \rho) \alpha}{\mu \beta} \mathcal{U}_v.
\end{equation}
Solving for $k^*$:
\begin{equation}
    k^* = \left( \frac{\varphi  (1 - \rho) \alpha}{\mu \beta} \right)^{\frac{1}{\mu+\varphi }} \cdot \mathcal{U}_v^{\frac{1}{\mu+\varphi }}.
\end{equation}
Let the scaling constant be $\kappa = (\frac{\varphi (1 - \rho) \alpha}{\mu \beta})^{\frac{1}{\mu+\varphi}}$. We finalize the Generalized Uncertainty-Driven Scaling Law:
\begin{equation}
    k^*(v) \propto \mathcal{U}_v^{\frac{1}{\mu+\varphi}}.
\end{equation}
This completes the proof. \qed

\section{More Implementation Details}\label{sec-app: parameter}

\subsection{Experimental Environment.}
We implement \model and all baselines using PyTorch 1.11, PyTorch Geometric 2.1.0, and Python 3.9.11 with CUDA 11.7. All experiments described in the main text are conducted on NVIDIA A100 GPUs (40GB memory). We utilize the NNI toolkit equipped with the Tree-structured Parzen Estimator (TPE) algorithm for automatic hyperparameter tuning.

\subsection{General Hyperparameter Search Space.}
For all methods (including baselines and \model), we employ a unified search space for the following general hyperparameters:
\begin{itemize}[leftmargin=10pt, nosep]
    \item \textbf{Learning rate:} \{0.005, 0.01, 0.05, 0.1\}
    \item \textbf{Weight decay:} \{5\text{e-}5, 1\text{e-}5, 1\text{e-}4, 5\text{e-}4, 5\text{e-}3\}
    \item \textbf{Dropout:} \{0.1, 0.3, 0.5, 0.7, 0.9\}
\end{itemize}

\subsection{Specific Hyperparameters for \model.}
In addition to the general parameters, we search for the specific hyperparameters of \model within the following spaces:
\begin{itemize}[leftmargin=10pt, nosep]
    \item \textbf{Load balancing coefficient ($\lambda_2$):} $\{1\text{e-}3, 1\text{e-}1\}$
    \item \textbf{Routing entropy coefficient ($\lambda_1$):} Fixed at $0.0001$
    \item \textbf{Sensitivity coefficient ($\gamma$):} $\{1, 5, 10\}$
    \item \textbf{Number of experts ($K$):} $\{4, 6\}$
    \item \textbf{Expert receptive field (hops):} $\{\text{``half\_half''}, \text{``all\_1hop''}\}$
    \item \textbf{Number of layers ($L$):} $\{1, 2\}$
    \item \textbf{Batch Normalization:} $\{\text{True}, \text{False}\}$
\end{itemize}

For the remaining baseline models, we adopt the hyperparameter search spaces as follows. Note that we adhere to the original notation definitions of these baselines; consequently, overlaps with the notation used in this paper are inevitable. However, they may carry different meanings, and we have provided symbol clarifications below.

\begin{itemize}[leftmargin=10pt, nosep]
  \item \textbf{FSGNN}: aggregation method for features from different hops (aggregator) $\in\{\text{cat}, \text{sum}\}$; 
  \item \textbf{FAGCN}: coefficient for preserving original node features ($\epsilon$)$ \in \{0.2,0.3,0.4,0.5\}$;
  \item \textbf{GPRGNN}: feature propagation steps ($K$) $\in \{10\}$, dropout $\in \{0.5\}$, PageRank restart probability ($\alpha$) $\in{0.5}$;
  \item \textbf{ACMGCN}: ACMGCN architecture variant selection (variant) $\in \{\text{False}\}$, whether structural information is needed (is\_need\_struct) $\in \{\text{False}\}$;
  \item \textbf{H2GCN}: number of neighborhood aggregation rounds (num\_layers) $\in \{1\}$, number of feature embedding MLP layers (num\_mlp\_layers) $\in \{1\}$;
  \item \textbf{ANS-GT}: number of data augmentations (data\_augmentation) $\in\{4,8,16,32\}$, number of model layers (n\_layer) $\in\{2,3,4\}$ and batch size $\in\{8,16,32\}$;  
  \item \textbf{NAGFormer}: hidden $\in\{128,256,512\}$, number of Transformer layers $\in\{1,2,3,4,5\}$ and number of propagation steps $\in\{7,10\}$;
  \item \textbf{SGFormer}: number of global attention layers is fixed as 1, number of GCN layers $\in\{1,2,3\}$, mixing weight for global semantics and local structure ($\alpha$) $\in\{0.5,0.8\}$;
  \item \textbf{Difformer}: hidden dimension $\in \{16,32,64\}$;
  \item \textbf{GMoE}: num\_layers $\in\{2,3\}$, load balancing loss coefficient (loss coef) $\in\{0.1,1\}$, num experts $\in\{4,8\}$, top $k\in\{1,2,4\}$, num experts 1hop $\in\{zero,half,all\}$;
  \item \textbf{NodeMoE}: Load balancing loss coefficient (balance) $\in\{0, 0.001, 0.01, 0.1, 1\}$, filter smoothing loss coefficient (gamma) $\in\{0, 0.01, 0.1, 1\}$, num experts $\in\{2, 3, 5\}$, dropout $\in\{0, 0.5, 0.8\}$;
  \item \textbf{DAMoE}: top $k\in\{2,3,4\}$, num experts $\in\{4,8\}$, load balancing loss coefficient (loss coef) $=0.001$, number of expert GNN layers (num\_layer) $=3$, minimum receptive field depth of the expert (min layer)=2;  
  \item \textbf{Mowst}: model1 $\in\{\text{MLP}\}$, model2 $\in\{\text{GCN, SAGE}\}$, model1\_num\_layers and model2\_num\_layers $\in\{2,3\}$ ;
  \item \textbf{Moscat}: learning rate of MLP gating $\in\{0.005, 0.001, 5\text{e-}4, 1\text{e-}4, 5\text{e-}5\}$, gating's normalization$\in\{\text{layer},\text{batch},\text{no norm}\}$, expert's maximum scope size ($L_{max}$) $\in\{6\}$.
\end{itemize}


\section{All Results for Ogbn-arxiv}

\begin{figure}[!htbp]
  \centering
  \includegraphics[width=\linewidth]{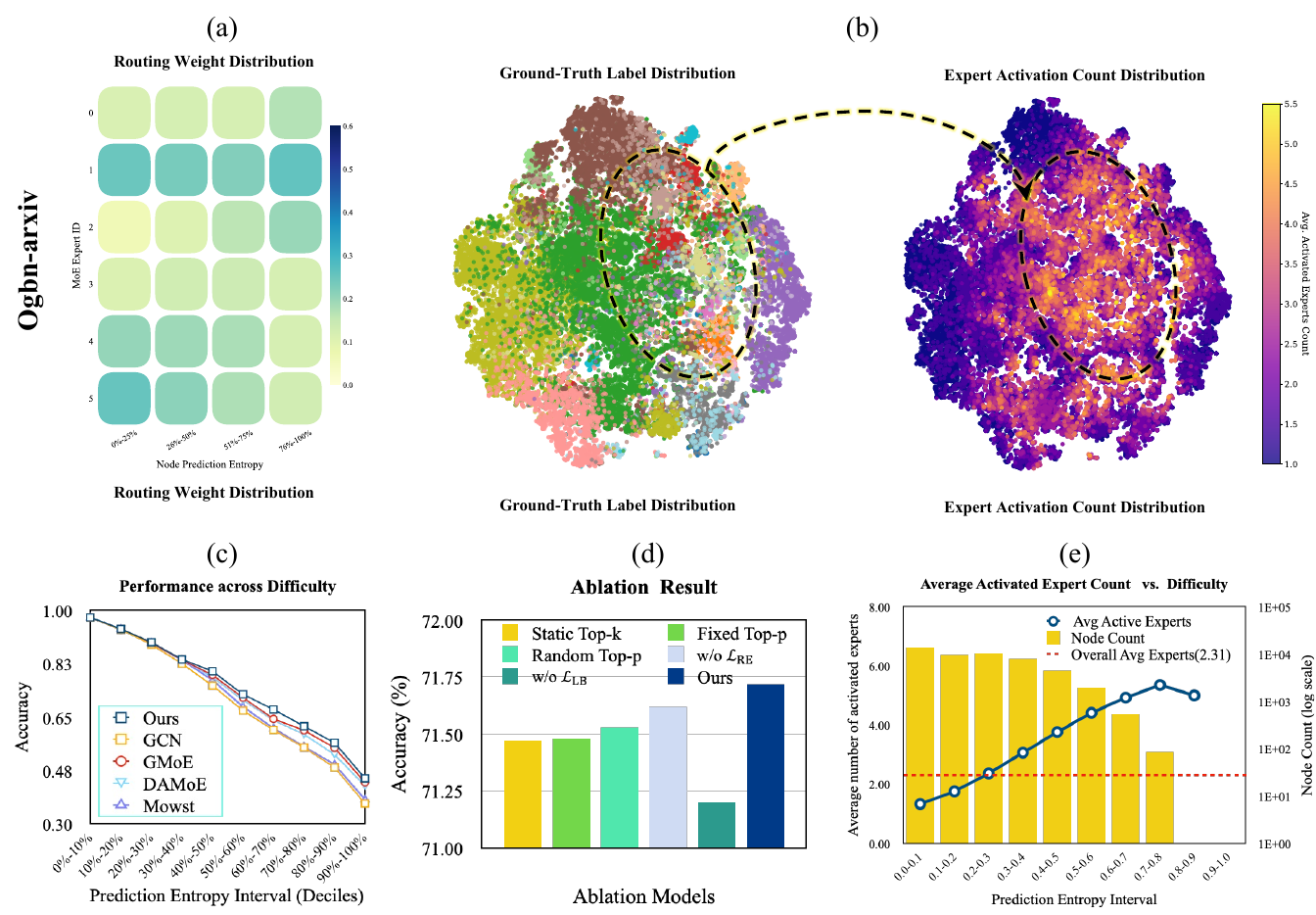} 
  \caption{All results for Ogbn-arxiv: 
  (a) Visualization of average routing weights;
  (b) Label distribution vs. expert activation count distribution;
  (c) Performance comparison across node difficulty levels;
  (d) Ablation analysis on \model;
  (e) Node density distribution and average number of acti-
  vated experts across varying levels of discriminative difficulty.}
  \label{fig-app: arxiv}
\end{figure}

\newpage

\end{document}